\documentclass[journal]{IEEEtran}

\usepackage{authblk}
\usepackage{graphicx}
\usepackage{amssymb,amsmath,epsfig}
\usepackage{algorithm}
\usepackage{algorithmic}
\usepackage{booktabs}
\usepackage{mathrsfs}
\usepackage{color}
\usepackage{textcomp}
\usepackage{bbding}
\usepackage{pifont}
\usepackage{wasysym}
\usepackage{amssymb}
\usepackage{subcaption}
\usepackage{cite}
\usepackage{amsmath,amssymb,amsfonts}
\usepackage{xcolor}
\usepackage{multirow}
\usepackage{epsfig}
\usepackage{colortbl}
\usepackage[colorlinks,
            linkcolor=red,
            anchorcolor=blue,
            citecolor=green
            ]{hyperref}

\begin{document}

\title{HAFormer: Unleashing the Power of Hierarchy-Aware Features for Lightweight \\ Semantic Segmentation}

\author{Guoan Xu, 
        Wenjing Jia,~\IEEEmembership{Member,~IEEE,} 
        Tao Wu, 
        Ligeng Chen, 
        and Guangwei Gao,~\IEEEmembership{Senior Member,~IEEE}

\thanks{This work was supported in part by the foundation of Key Laboratory of Artificial Intelligence of Ministry of Education under Grant AI202404, in part by the Open Fund Project of Provincial Key Laboratory for Computer Information Processing Technology (Soochow University) under Grant KJS2274.~\textit{(Corresponding author: Wenjing Jia; Guangwei Gao.)}}

\thanks{Guoan Xu and Wenjing Jia are with the Faculty of Engineering and Information Technology, University of Technology Sydney, Sydney, NSW 2007, Australia (e-mail: xga\_njupt@163.com, Wenjing.Jia@uts.edu.au).}

\thanks{Tao wu is with the State Key Laboratory of Novel Software Technology, Nanjing University, Nanjing 210023, China (e-mail: wt@smail.nju.edu.cn).}

\thanks{Ligeng Chen is with the State Key Laboratory of Novel Software Technology, Nanjing University, Nanjing 210023, China, and also with Honor Device Company, Ltd, Shenzhen 518040, China(e-mail: chenlg@smail.nju.edu.cn).}

\thanks{Guangwei Gao is with the Institute of Advanced Technology, Nanjing University of Posts and Telecommunications, Nanjing 210023, China, and also with the Key Laboratory of Artificial Intelligence, Ministry of Education, Shanghai 200240, China (e-mail: csggao@gmail.com).}
}

\markboth{IEEE Transactions on Image Processing}%
{Shell \MakeLowercase{\textit{et al.}}: Bare Demo of IEEEtran.cls for IEEE Journals}

\maketitle

\begin{abstract}
Both Convolutional Neural Networks (CNNs) and Transformers have shown great success in semantic segmentation tasks. Efforts have been made to integrate CNNs with Transformer models to capture both local and global context interactions. However, there is still room for enhancement, particularly when considering constraints on computational resources. In this paper, we introduce HAFormer, a model that combines the hierarchical features extraction ability of CNNs with the global dependency modeling capability of Transformers to tackle lightweight semantic segmentation challenges. Specifically, we design a Hierarchy-Aware Pixel-Excitation (HAPE) module for adaptive multi-scale local feature extraction. During the global perception modeling, we devise an Efficient Transformer (ET) module streamlining the quadratic calculations associated with traditional Transformers. Moreover, a correlation-weighted Fusion (cwF) module selectively merges diverse feature representations, significantly enhancing predictive accuracy. HAFormer achieves high performance with minimal computational overhead and compact model size, achieving 74.2\% mIoU on Cityscapes and 71.1\% mIoU on CamVid test datasets, with frame rates of 105FPS and 118FPS on a single 2080Ti GPU. The source codes are available at \textit{https://github.com/XU-GITHUB-curry/HAFormer}.

\end{abstract}


\IEEEpeerreviewmaketitle

\section{Introduction}
\label{sec1}

\IEEEPARstart{S}{emantic} segmentation involves the task of assigning a label to each pixel in a given image, making it a fundamental dense prediction task in computer vision with applications in autonomous driving~\cite{hu2023planning}, medical care~\cite{yuan2023effective}, satellite remote sensing~\cite{wang2023geometric}, and more. 
Previous methods, such as~\cite{fu2019dual,li2023context}, leverage deep convolutional neural networks (CNNs) for feature extraction, incorporating feature pyramid structures for multi-scale information perception~\cite{yang2018denseaspp} and attention modules for global context perception~\cite{woo2018cbam, wang2018non, zhu2019asymmetric}. 
Although these methods have achieved considerable accuracy, they often 
require extensive computational resources and exhibit relatively slow inference speeds due to deep network stacking for larger receptive fields and higher semantic levels.

\begin{figure}[t]
	\centerline{\includegraphics[width=8cm]{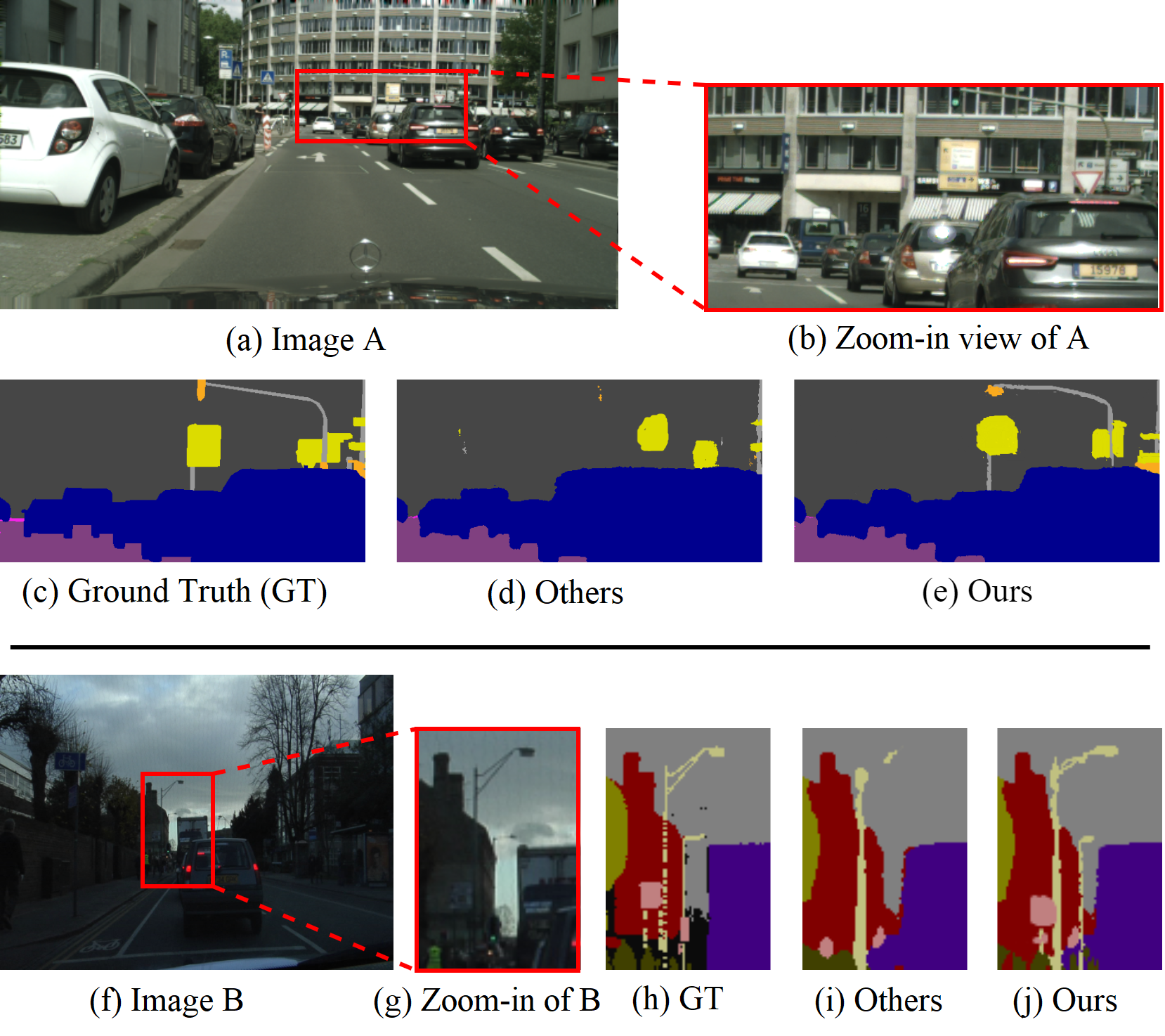}}
         \caption{Visual comparison of small object segmentation using our approach versus an existing method on sample images from Cityscapes (top) and CamVid (bottom).}
	\label{fig:smallvs}
\end{figure}

To accommodate devices with limited computational resources, recent studies~\cite{paszke2016enet,romera2017erfnet,mehta2018espnet, zhao2018icnet,gao2021mscfnet} have focused on developing lightweight segmentation models. 
For example, ERFNet~\cite{romera2017erfnet} employs 1-D non-bottleneck modules to reduce computation, while ICNet~\cite{zhao2018icnet} utilizes inputs of varying resolutions to enhance information flow across different branches. FBSNet~\cite{gao2022fbsnet} uses a symmetrical encoder-decoder structure with a spatial detail branch and a semantic information branch to refine 
contextual details. 
Typically, these models simplify the base module structures to minimize computational costs. However, while enhancing computational efficiency, their segmentation accuracy is often compromised due to the local limitations of convolution networks and shallower network depths.

Transformers have recently demonstrated remarkable success in various computer vision communities~\cite{zheng2021rethinking,liu2021swin}. Drawing inspiration from this progress, researchers have started integrating ViT~\cite{dosovitskiy2020image} architectures to tackle semantic segmentation challenges. Unlike CNNs, Transformers inherently provide a broad global receptive field through their extensive global attention mechanisms. Models using Transformers as image encoders excel in global context modeling, leading to significant improvements in segmentation accuracy compared to CNN-based approaches. While UNETR~\cite{hatamizadeh2022unetr} and other methods~\cite{wu2023d, heidari2023hiformer} base predictions on the last layer of the Transformer encoder, they tend to overlook smaller-scale objects in images, affecting the precise classification of smaller elements or pixels, as depicted in Fig.~\ref{fig:smallvs}. 
SegFormer~\cite{xie2021segformer} introduces a hierarchical attention model integrating a hierarchical Transformer encoder and a lightweight multi-layer perceptron (MLP) decoder to enhance segmentation precision. MPViT~\cite{lee2022mpvit} effectively incorporates multi-scale feature inputs into Transformer operations, yielding impressive results. 

These methods prioritize 
high segmentation accuracy but often overlook model efficiency. Firstly, transformer-based approaches lack inductive bias, making their training slow and challenging to converge. In addition, they typically require larger datasets and extended training duration, resulting in significant training overhead. Secondly, slow inference speeds are attributed to the time-consuming 
multi-head self-attention (MHSA) operations. The computational burden escalates, especially with high-resolution inputs, due to 
the quadratic complexity of MHSA. Additionally, these methods may struggle with capturing details and small objects due to their limited fine local modeling capabilities. 

In this work, our goal is to develop a lightweight semantic segmentation model that leverages both CNNs and Transformers, focusing on minimizing model size and computational requirements. Introducing the ``HAFormer" model, we combine the global receptive capabilities of Transformers with the local perception strengths of CNNs to unleash the power of hierarchy-aware features.

The core contributions of this paper are threefold:
\begin{itemize}
\item We propose a novel 
Hierarchy-Aware Pixel-Excitation (HAPE) module, utilizing hierarchy and content-aware attention mechanisms to reduce the computational load while enabling the extraction of deeper semantic information from pixels under various receptive fields. 

\item We develop an effective feature fusion mechanism, named correlation-weighted Fusion (cwF), to synergistically integrate the local and global context features learned by CNNs and Transformers, effectively enhancing accuracy. 
\item We propose an Efficient Transformer to decompose $Q$, $K$, and $V$ matrices, which effectively addresses the quadratic computational complexity challenge present in traditional Transformer models. 
\end{itemize}
Extensive experiments conducted on two widely used benchmarks demonstrate that our HAFormer achieves a balance between segmentation accuracy and efficiency.

The remainder of this paper is structured as follows: 
Section~\ref{sec2} provides a comprehensive review of related works. 
Section~\ref{sec3} presents the 
details of our proposed HAFormer, focusing on 
its three key components. 
Section~\ref{sec4} 
describes the detailed experimental setting and presents the 
evaluation results, including 
ablation studies and discussions. 
Finally, 
Section~\ref{sec5} concludes the paper by summarizing the key findings and discussing future directions.

\section{Related Work}
\label{sec2}


\subsection{Hierarchical Methods in Semantic Segmentation}
\label{sec21}

In dense prediction tasks, 
accurately classifying multi-scale and small target objects is a common challenge. 
This is particularly evident in semantic segmentation, where the classification of small objects can be affected by neighboring larger objects, leading to misclassification. 
Hierarchical methods effectively address this challenge by utilizing convolutions with varying dilation 
rates or pooling layers with different rates. 
The outcomes are then cascaded or concatenated to integrate information from diverse scales. This multi-scale integration enhances receptive field levels, mitigating ambiguity from varying local region sizes and improving object detail handling. 
Existing hierarchical approaches~\cite{zhao2018icnet,yuan2021hrformer,li2019dfanet, gao2021mscfnet, xie2021segformer,yang2018denseaspp} can be classified into overall hierarchical structures or specific hierarchical modules, summarized as follows:

\textbf{Hierarchical Structures.} 
Several approaches have adopted a multi-scale design, featuring distinct network branches handling inputs or feature maps of varying resolutions. 
A notable method following this approach is ICNet~\cite{zhao2018icnet}, which incorporates three encoding branches (low-resolution, medium-resolution, and high-resolution), each 
excelling at extracting fine-grained information at different scales to enhance boundary information in the output. 
In contrast, HRFormer~\cite{yuan2021hrformer} effectively combines robust semantic information with precise location details. 
Whereas 
HSSN~\cite{li2022deep} is a hierarchical approach, focusing 
on categorizing objects like ``Human-Rider-Bicyclist'' rather than addressing pixel-level classification challenges for small objects. Other methods, including~\cite{li2019dfanet, gao2021mscfnet, xie2021segformer,li2022mvitv2}, utilize multi-scale structures by parallelizing multiple resolution branches and facilitating continuous information interaction among them. 

\textbf{Hierarchical Modules.} 
Numerous methods integrate hierarchical modules at specific layers within the architecture, allowing the utilization of varied receptive fields on feature maps. 
For example, the ASPP module used in DeepLab~\cite{chen2017deeplab, chen2018encoder} and DenseASPP~\cite{yang2018denseaspp} effectively extracts features from different scales through atrous convolutions, addressing the variability in object scales within and across images. 
PSPNet~\cite{zhao2017pyramid} stands out for its pyramid pooling module that integrates features from four scales. By collecting and merging contextual information from diverse scales, this module generates more representative and discriminative features than those from global pooling alone. 
Models using this module can enhance their recognition capability for objects of various sizes. 
Inspired by the ``wider" modules~\cite{zhao2017pyramid, wang2022pvt},  
in this work we demonstrate that utilizing multiple diverse convolution kernels efficiently enhances expressive capacity, leading to improved performance with minimal computational and parameter overhead.

\begin{figure*}[htbp]
	\centerline{\includegraphics[width=17.5cm]{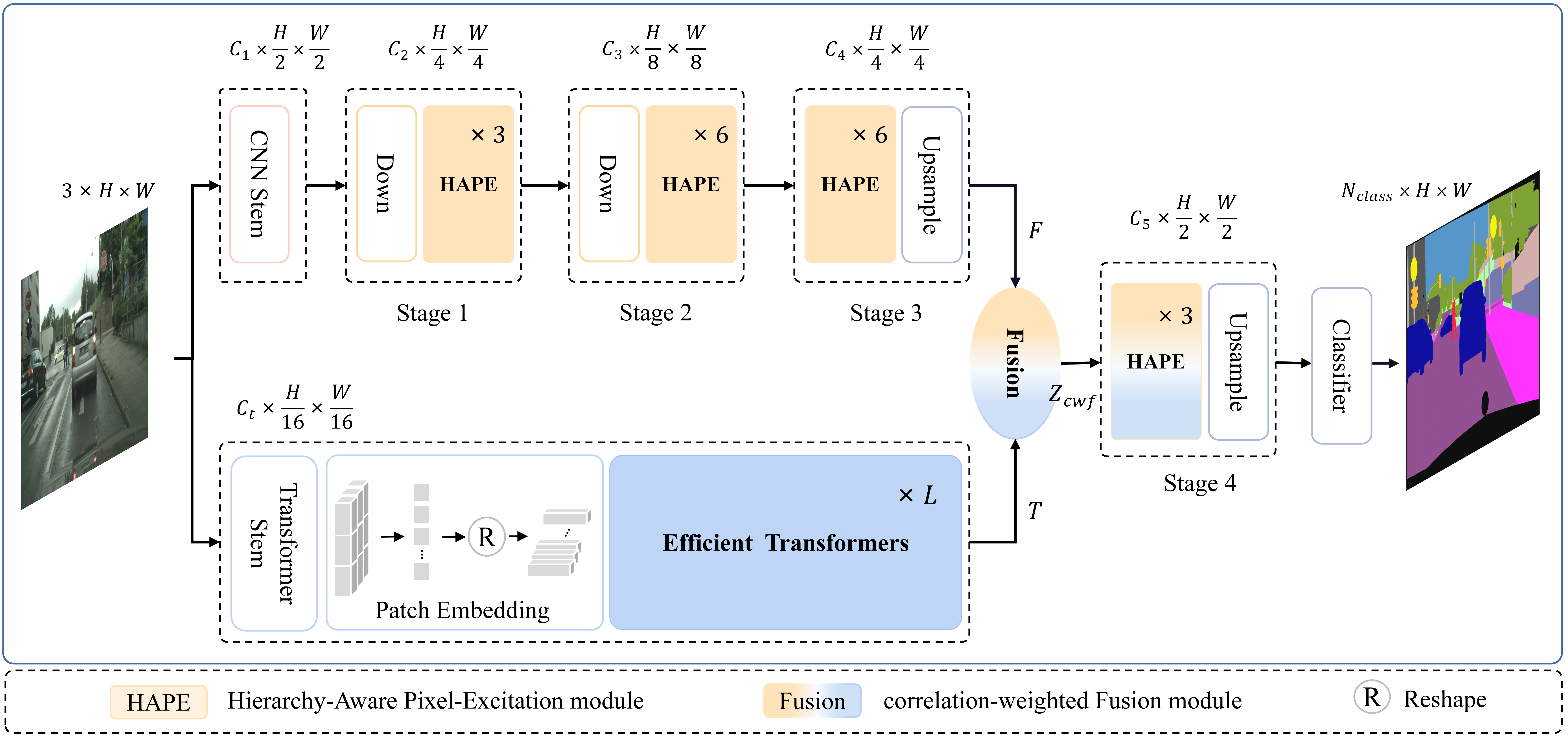}}
	\caption{The overall architecture 
 of the proposed HAFormer. HAFormer introduces a Hierarchy-Aware Pixel-Excitation (HAPE) module for adaptive multi-scale local feature extraction. For global perception modeling, HAFormer develops an efficient Transformer module to streamline the quadratic calculations. Additionally, a correlation-weighted Fusion (cwF) module selectively combines diverse feature representations, markedly boosting predictive accuracy.}
	\label{fig:net}
\end{figure*}

\subsection{Vision Transformer in Semantic Segmentation}
\label{sec22}

The groundbreaking ViT~\cite{dosovitskiy2020image} introduces a pure transformer framework for image recognition, treating images as sequences of patches processed through multiple layers. 
Subsequent models such as 
DeiT~\cite{touvron2021training}, Fact~\cite{jie2023fact}, CrossFormer~\cite{wang2023crossformer++}, and DViT~\cite{yao2023dual} have further excelled in image processing tasks. 
SETR~\cite{zheng2021rethinking} is a tailored paradigm for segmentation, utilizing a pure Transformer model in the encoder and various CNN decoder combinations to achieve state-of-the-art results. 
Swin-Transformer~\cite{liu2021swin} addresses redundant computations, easing computational loads to some extent. 
However, these methods still require extensive training data to match CNN performance, posing challenges in dense prediction fields requiring detailed annotations. 
Transformer-based models such as
~\cite{lee2022mpvit, li2022mvitv2} have recognized the importance of hierarchical perceptions in dense prediction tasks, incorporating multi-scale structures and pyramid modules in their designs. 

Recent studies have noted that Transformers often prioritize global long-range dependencies, potentially overlooking critical features like local connections and translation invariance characteristic of CNNs. 
Consequently, various methods~\cite{fan2023segtransconv, wang2022uctransnet, yuan2023effective, gao2023ctcnet} have sought to combine 
CNNs and Transformers to fully leverage the strengths of both. 
However, these efforts struggle to balance real-time inference requirements and low-latency capabilities. Lightweight techniques such as 
LETNet~\cite{xu2023lightweight} position the Transformer as a capsule network 
while others, such as TopFormer~\cite{zhang2022topformer}, integrate it as an auxiliary component in the decoder to enhance boundary recovery. 
Nonetheless, a definitive solution for effectively combining global and local information remains elusive. 

To tackle the challenges of high computational requirements and effectively integrating local information with a global context when combining CNNs with Transformers, our HAFormer introduces an Efficient Transformer (ET) module to manage computational complexity and a correlation-weighted Fusion (cwF) mechanism to harmonize features from CNNs and Transformers.

\subsection{Attention Mechanisms in Semantic Segmentation}
\label{sec23}

Inspired by the focal nature of human visual perception, attention mechanisms emphasize significant features while disregarding irrelevant ones. These mechanisms fall into two main categories: channel attention and spatial attention. 
In channel attention methods, SKNet~\cite{li2019selective} enables neurons to dynamically adjust their receptive field sizes based on input scales. Spatial attention methods, such as non-local neural networks~\cite{wang2018non}, capture long-range dependencies in semantic segmentation. However, modeling relationships between all locations can be computationally intensive. Asymmetric non-local neural networks~\cite{zhu2019asymmetric} attempted to reduce computational costs, yet they may still be resource-intensive, especially with high-resolution input features. 

Researchers have explored combining both channel and spatial attention mechanisms to enhance features from multiple perspectives. 
For instance, CBAM~\cite{woo2018cbam} sequentially operates along two independent dimensions (channel and spatial), producing attention maps that are then multiplied with input features for adaptive feature optimization. DANet~\cite{fu2019dual} and CCNet~\cite{huang2019ccnet} integrate channel and spatial attention in parallel, employing self-attention operations and combining the resulting features. 
CAA~\cite{huang2021channelized} disassembles axial attention and integrated channel attention to manage conflicts and prioritize features. 
These methods, utilizing self-attention mechanisms, have demonstrated positive results. 

A prevalent challenge involves pixel-wise long-distance modeling, which incurs high computational costs, rendering it unsuitable for deployment in resource-constrained scenarios. 
This study introduces a lightweight model that optimizes the local perception of CNNs and the global modeling abilities of Transformers. 
We address the computational complexity issue by utilizing a spatial reduction-linear projection and split operation strategy within our proposed Efficient Transformer (ET) module. 

\section{The Proposed Method}
\label{sec3}


\subsection{Overall Architecture}
\label{sec31}

The overall architecture of our HAFormer is illustrated in Fig.~\ref{fig:net}, which features three components: a CNN encoder enhanced with hierarchy-aware pixel excitation, an efficient Transformer encoder, and a lightweight decoder. 

For a given input image ${I \in {\mathbb{R}^{3  \times H \times W}}}$ with dimensions of ${H \times W}$, the model begins with a CNN Encoder, producing features ${F \in {\mathbb{R}^{{C_f } \times H_f \times W_f}}}$$\left(H_f = \frac{H}{8}, W_f = \frac{W}{8}\right)$. 
Simultaneously, the input $I$ undergoes processing in the Transformer encoder post the Transformer Stem block, resulting in feature embedding ${T \in {\mathbb{R}^{N \times D}}}$, where ${N = \frac{{{H_t}}}{P} \times \frac{{{W_t}}}{P}}$$\left(H_t = \frac{H}{16}, W_t = \frac{W}{16}\right)$ denotes the token count, ${D = {C_t} \times {P^2} }$ denotes the dimension of each token, and $P$ signifies token size. 
Subsequently, the two distinct types of context features, $F$ and $T$, are synergized effectively by our newly designed correlation-weighted Fusion (cwF) module. This fusion of correlated CNN and Transformer features enhances boundary information and restoration with the lightweight decoder segmentation head. 



Specifically, to optimize the CNN encoder, we employ three $3 \times 3$ convolutional layers in the CNN Stem block. 
In this configuration, the last layer has a stride of 2, resulting in a feature map size of ${{C_1} \times \frac{H}{2} \times \frac{W}{2}}$, where $C_1$ denotes the output channel count. 
In contrast, the Transformer stem in the Transformer Encoder reduces the resolution while extracting feature representations, contributing to the model's lightweight design by minimizing computational load, since higher resolution means more computation. 
Therefore, in the Transformer Stem block, we employ four $3 \times 3$ convolution layers with a stride of 2, resulting in an output feature size of ${{C_t } \times \frac{H}{{16}} \times \frac{W}{{16}}}$. 


\subsection{Hierarchy-Aware Pixel-Excitation (HAPE) Module}
\label{sec32}

Employing convolutions with diverse kernel sizes within the same layer, combined with pixel excitation, facilitates feature extraction from objects of varying sizes. 
Building on this concept and drawing inspiration from works like~\cite{zhao2017pyramid, yang2023cswin}, we adopt a multi-scale strategy to capture unique pixel features across different receptive field levels. Unlike the layer-wise merging seen in ESPNet~\cite{mehta2018espnet} and concatenation in Inception~\cite{szegedy2016rethinking}, our module avoids redundant computations, leading to a more streamlined network while preserving feature effectiveness. 
Additionally, to further improve pixel representation across diverse scales, we introduce the innovative Hierarchy-Aware Pixel Excitation (HAPE) module in this study. 
This module enhances the model's ability to effectively recognize objects of various sizes in an image, ultimately reducing pixel misclassification rates. 

\begin{figure}[t]
	\centerline{\includegraphics[width=1.0\linewidth]{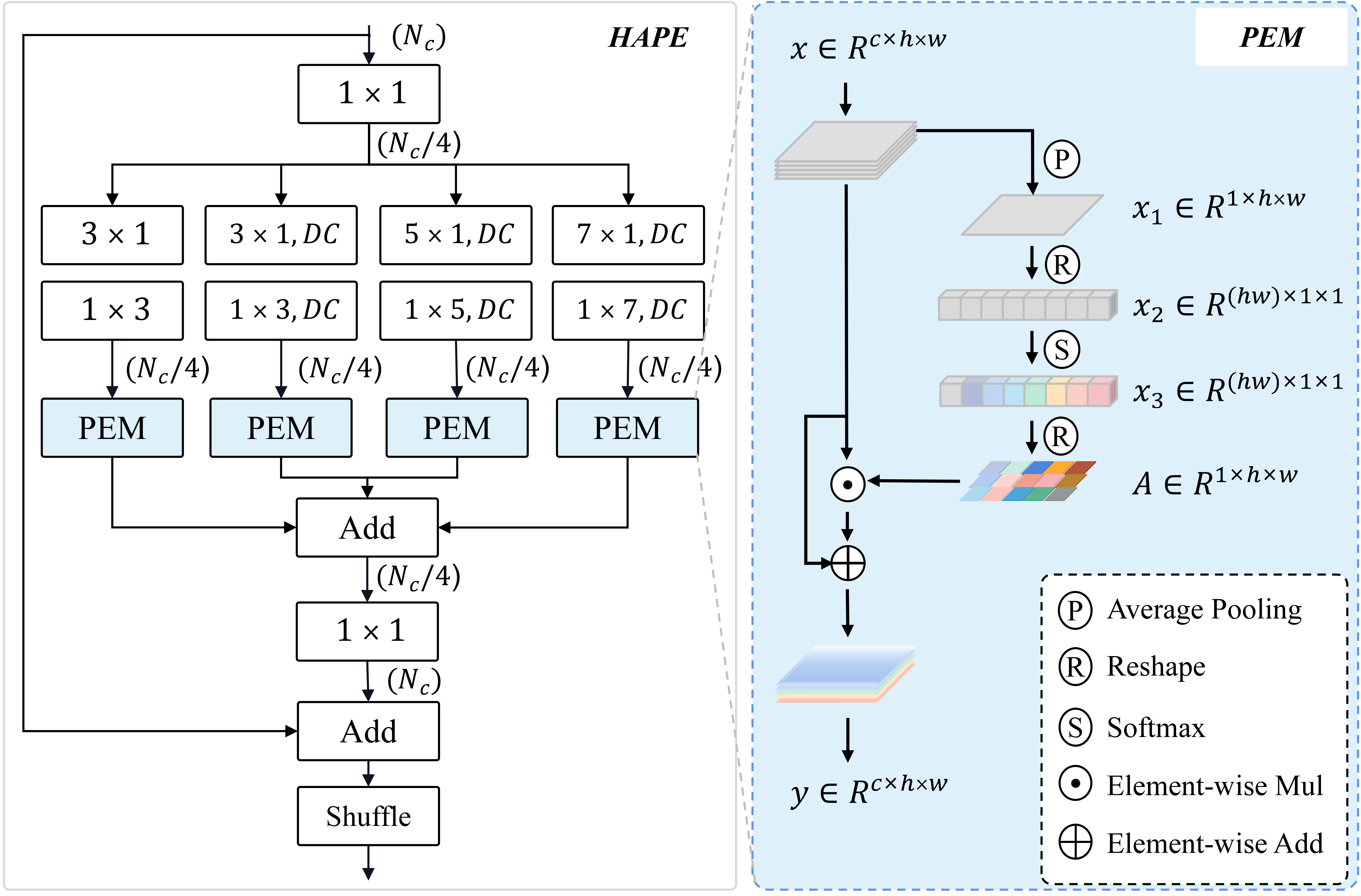}} 
	\caption{The architecture of our Hierarchy-Aware Pixel-Excitation (HAPE). $DC$ stands for dilation convolution.}
	\label{fig:hape}
\end{figure}

Specifically, as depicted in Fig.~\ref{fig:hape}, given a feature input $ {X_{in}} \in {\mathbb{R}^{N_c \times H_c \times W_c }} $, we initially feed it into a $1 \times 1$ convolutional layer to reduce its channel dimensions to $\frac{N_c}{4}$, \textit{i.e.}, the output feature map $ \mathop X\limits^ \sim $ is denoted as 
\begin{equation}
    \mathop X\limits^ \sim   = {f_{1 \times 1}}\left( {{X_{in}}} \right), \mathop X\limits^ \sim   \in {\mathbb{R}^{\frac{N_c}{4} \times H_c \times W_c }}.  
\end{equation}
Here, $f_{1 \times 1}$ denotes a convolution operation with a kernel size of $1 \times1 $. This dimension reduction facilitates the channel operation in the subsequent hierarchical convolutional layers. 

\begin{figure}[t]
	\centerline{\includegraphics[width = \linewidth]{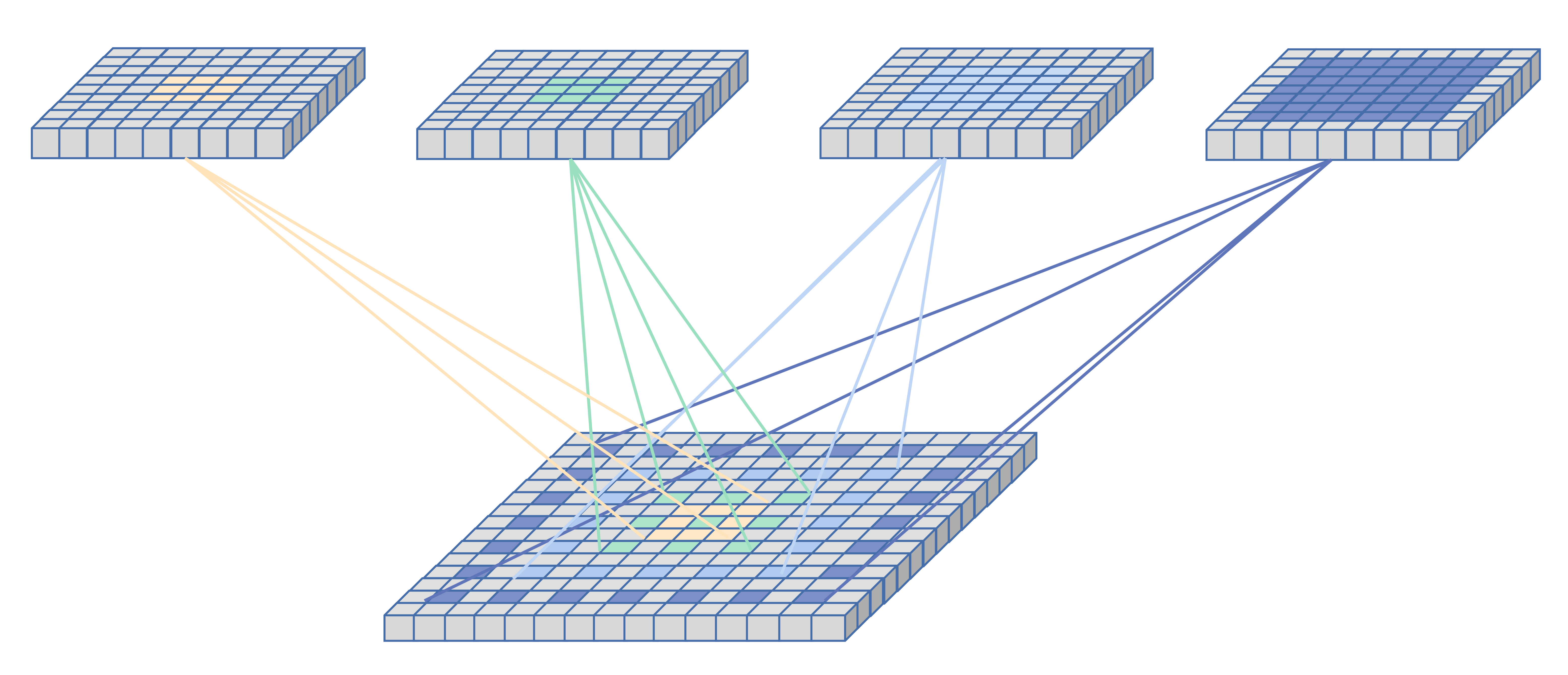}}
	\caption{Supplementary instructions of hierarchical respective 
 fields for HAPE.}
	\label{fig:hier}
\end{figure}

Subsequently, we perform four parallel convolution operations, comprising factorized convolution and depthwise separable convolution, with kernel sizes of 3, 3, 5, and 7, respectively. Additionally, the last three convolutional layers utilize dilated convolution to enhance receptive fields, as shown in Fig.~\ref{fig:hier}. This strategy enables the model to capture image features across various scales, ensuring comprehensive and detailed information extraction. 

The above process is expressed as 
\begin{equation}
    {l_1} = {f_{1 \times 3}}\left( {{f_{3 \times 1}}\left( {\mathop X\limits^ \sim  } \right)} \right),
\end{equation}
\begin{equation}
   {l_i} = f_{1 \times {k_i}}^{d_c}\left( {f_{{k_i} \times 1}^{d_c}\left( {\mathop X\limits^ \sim  } \right)} \right),\left\{ {{k_i} = 3,5,7;i = 2,3,4} \right\}, 
\end{equation}
where ${l_i}$ represents the intermediate features, $f_{1 \times {k_i}}$ is a 1-D convolution operation with a kernel size of ${k_i}$, and $d_c$ denotes the dilation rate. 
For simplicity, 
some activation and batch normalization operations are excluded from the equations.

\begin{figure*}[!t]
	\centerline{\includegraphics[width=0.8\linewidth]{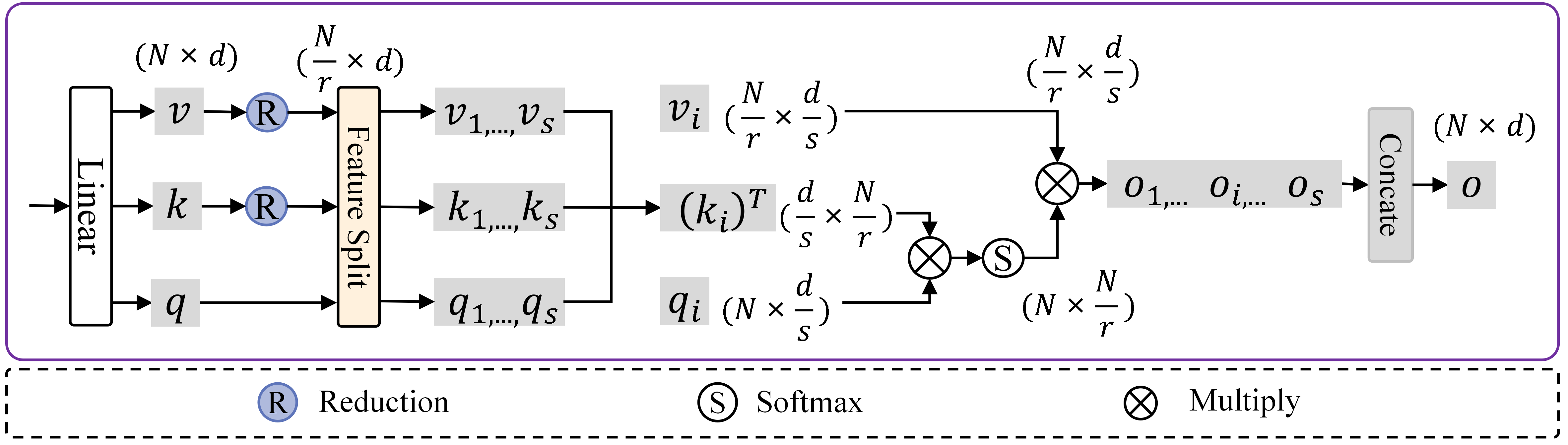}}
	\caption{The architecture of the proposed efficient Multi-Head Self-Attention (eMHSA).}
	\label{fig:EMHSA}
\end{figure*}

\begin{figure*}[t]
	\centerline{\includegraphics[width=0.7\linewidth]{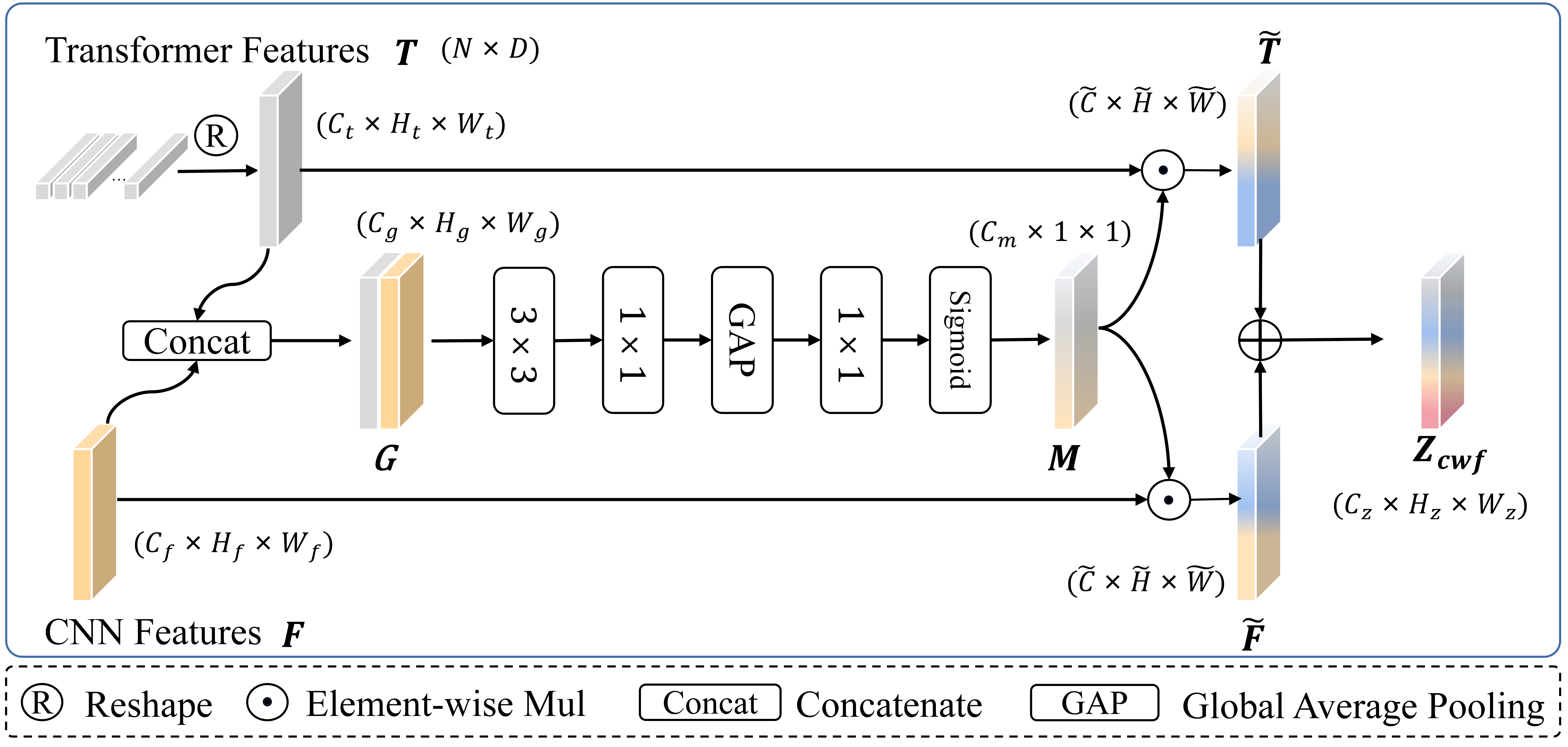}}
	\caption{The architecture of the correlation-weighted Fusion (cwF) module.}
	\label{fig:cwf}
\end{figure*}

A critical element lies in the Pixel-Excitation Module (PEM), which is responsible for enhancing the feature representability through a content-aware spatial-attention mechanism. 
As illustrated 
in Fig.~\ref{fig:hape}, the process begins by feeding the input ${x \in {\mathbb{R}^{c \times h \times w}}}$ into the Global Average Pooling (GAP) layer, generating ${{x_1} \in {\mathbb{R}^{1 \times h \times w}}}$. Subsequently,  ${x_1}$ undergoes reshaping 
and flattening operations before being input to the $Softmax$ function to calculate the weight matrix ${A \in {\mathbb{R}^{1 \times h \times w}}}$. 
This weight matrix is then multiplied with the input features, resulting in a content-aware attention-enhanced output $x^{\prime}$. 

This process can be represented as
\begin{equation}
    {x_1} = \text{Reshape}\left( {\text{GAP}\left( x \right)} \right),
\end{equation}
\begin{equation}
    A = {\text{Reshape}^{ - 1}}\left( {\text{Softmax}\left( {{x_1}} \right)} \right),
\end{equation}
and
\begin{equation}
    x^{\prime} = \delta \left( {x \odot A + x} \right).
\end{equation}
Here, $Reshape$ and $Reshape^{ - 1}$ denote the reshaping 
operation and its reverse operation,  
$\delta$ is an activation function, and $\odot$ denotes 
element-wise multiplication.

Finally, a residual structure is employed to retain the original features, yielding the final output ${Y \in {\mathbb{R}^{N_c \times H_c \times W_c}}}$. The four convolution layers are jointly added into a $1 \times 1$ convolution for feature fusion and channel restoration. A residual connection is maintained within the module, and the channel shuffle operation effectively facilitates the information interaction between channels, as expressed as
\begin{equation}
Y = \text{Shuffle}\left( {{f_{1 \times 1}}\left( {\delta \left( {\sum\limits_{i = 1}^4 {\text{PEM}\left( {{l_i}} \right)} } \right)} \right) + X_{in}} \right),
\end{equation}
where $Shuffle$ represents 
the channel shuffle operation, and $\delta$ is an activation function.

\begin{figure*}[t]
	\centerline{\includegraphics[width=0.98\linewidth]{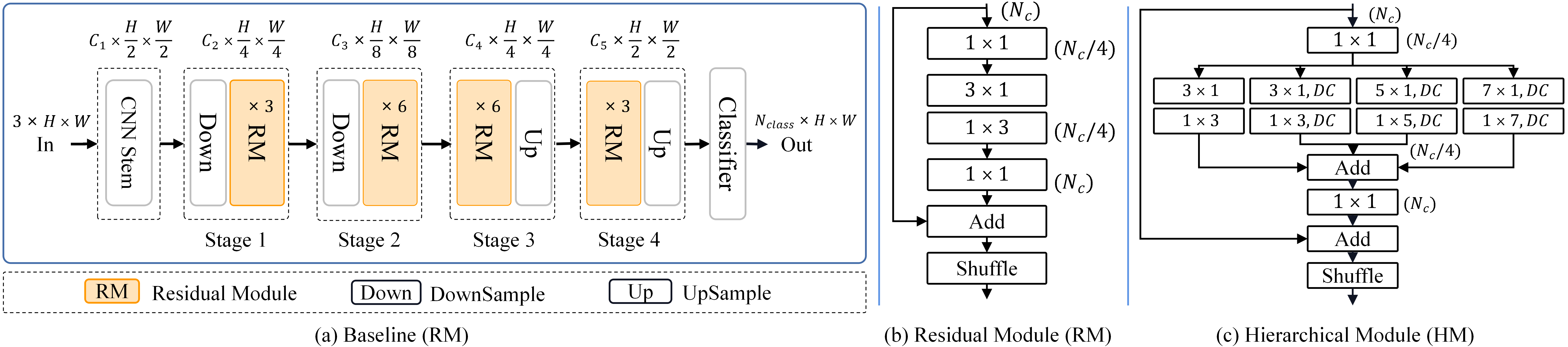}}
	\caption{The architecture of the baseline model (a), the detailed illustration of the Residual Module (RM)  used in the baseline (b), and the Hierarchical Module (HM) in our proposed HAPE module (c).}
	\label{fig:base}
\end{figure*}

\subsection{Efficient Transformer}
\label{sec33}

Conventional Transformer methods, as evidenced by~\cite{liu2021swin,wu2021cvt}, can be excessively large for lightweight and real-time models, especially when handling high-resolution inputs. 
This underscores the urgent need for more efficient Transformers. 
Inspired by~\cite{lu2022transformer, wang2022pvt}, our approach focuses on reducing computational costs by diminishing feature dimensions without significant loss of image details. 
To achieve this, we introduce a spatial reduction linear projection method which 
initially maps features into a latent embedding space with reduced dimensions before employing them for multi-head self-attention calculations. 
This approach, known as efficient Multi-Head Self-Attention (eMHSA) with learned projection and split operation, is depicted in Fig.~\ref{fig:EMHSA}. 


Denote the input feature as $X_t \in {\mathbb{R}^{C_t \times H_t \times W_t}}$, where $C_t$, $H_t$, and $W_t$ represent the number of the channel, height, and width of the feature map, respectively. Following the $Reshape$ operation, a sequence of flattened non-overlapping patches is derived, resulting in $X_t \in {\mathbb{R}^{N \times \left( {C_t \cdot {P^2}} \right)}}$, where $N = \frac{{{H_t}{W_t}}}{{{P^2}}}$ indicates the number of patches (\textit{i.e.}, the input sequence length), with each patch size being $P \times P$. Subsequently, the patches are mapped by a learnable linear projection layer $E \in {\mathbb{R}^{\left( {{P^2} \cdot C_t} \right) \times D}}$ into a latent $D$ dimensional embedding space, denoted as $Z \in {\mathbb{R}^{N \times D}}$. This process can be formulated as 
\begin{equation}
     Z = \left[ {x_p^1E;x_p^2E;...;x_p^NE} \right],
\end{equation}
where $x_p^i$ denotes the $i$-th patch. 
Note that the omission of position embedding is intentional to allow greater adaptability for different input sizes.

Subsequently, the three matrices in Transformers, namely the queries $Q$, the keys $K$, and the values $V$ are derived through their linear projections $W^Q$, $W^K$, and ${W^V} \in {\mathbb{R}^{D \times {D_h}}}$. This can be expressed as
\begin{equation}
     Q,K,V = Z{W^Q},Z{W^K},Z{W^V} \in {\mathbb{R}^{N \times {D_h}}}.
\end{equation}

Moreover, the number of heads $h$ in the multi-head self-attention is also a user-defined parameter, ensuring each head's dimension equals $d = \frac{{{D_h}}}{h}$. 
Consequently, the dimensions of ${q}$, ${k}$, and ${v}$ in the $i$-th head are $N \times d$. 
In the $i$-th head, $k$ and $v$ undergo spatial reduction by a factor of $r$, where $r$ is the reduction ratio and set to 2. 
Then, the sub-tokens resulting from the feature split operation undergo matrix multiplication with a field representing only $\frac{1}{s}$ of the original perception, where $s$ denotes the number of feature splits, which is set to 4. 
This process can be described as
\begin{equation}
\left( {{q_1}_{,...,}{q_s}} \right),\left( {{k_1}_{,...,}{k_s}} \right),\left( {{v_1}_{,...,}{v_s}} \right) = \text{Feature}\_\text{Split}\left( {q,k,v} \right).
\end{equation}

Therefore, the spatial distribution becomes ${q_i} \in {\mathbb{R}^{N \times \frac{d}{s}}}$, ${k_i} \in {\mathbb{R}^{\frac{N}{{{r}}} \times \frac{d}{s}}}$, and ${v_i} \in {\mathbb{R}^{\frac{N}{{{r}}} \times \frac{d}{s}}}$. This idea shares similarities with the concept of group convolution and can efficiently reduce memory consumption. Thus, the self-attention in the $n$-th head is calculated as
\begin{equation}
{o_i}\left( {{q_i},{k_i},{v_i}} \right) = \text{Softmax}\left( {\frac{{{q_i}{{\left( {{k_i}} \right)}^T}}}{{\sqrt d }}} \right){v_i},i \in \left[ {1,s} \right],
\end{equation}
and
\begin{equation}
hea{d^n} = \text{Concat}[{o_1},{o_2},...,{o_s}],n \in \left[ {1,h} \right],
\end{equation}
where $Concat[.,. ]$ denotes the concatenating operation.

Thus, the final output of the eMHSA is denoted as 
\begin{equation}
\text{eMHSA} = \text{Concat}[hea{d^1},hea{d^2},...,hea{d^h}]{W^O},
\end{equation}
where $h$ represents the number of heads in eMHSA, while ${W^O} \in {\mathbb{R}^{{D_h} \times D}}$ serves as a linear projection to restore the dimension. Hence, with the structure designed above, we have reduced the complexity from $O\left( {{N^2}} \right)$ to $O\left( {\frac{{{N^2}}}{{s{r}}}} \right)$.

It is noteworthy that the Transformer series~\cite{zheng2021rethinking, liu2021swin, lee2022mpvit} also utilize a kind of self-attention mechanism, including Multi-Head. However, their approach 
is computationally 
intensive for capturing detailed relationships among features, which deviates from our objectives. 

As for the MLP layer, we follow the approach described in~\cite{raghu2021vision, wang2022pvt}, replacing fixed-size position encoding with zero-padding position encoding. Moreover, we introduce a depth-wise convolution with a padding size of 1 to capture local continuity in the input tensor between the fully connected (FC) layer and GELU in the feed-forward networks. By eliminating fixed-size positional embeddings, the model becomes versatile in handling inputs with different resolutions. Thus, the output of the efficient MLP layer, denoted as ``$eMLP$", can be written as
\begin{equation}
\text{eMLP} = \rho \left( {{\xi _{\text{GELU}}}\left( {{f_{\text{DWConv}}}\left( {\rho \left( {{x_{e}}} \right)} \right)} \right)} \right),
\end{equation}
where $\rho$ denotes the FC layer operation, ${\xi _{\text{GELU}}}$ represents the GELU activation function, ${f_{\text{DWConv}}}$ signifies depthwise convolution, and $x_{e}$ is the input of eMLP.

\subsection{ Correlation-weighted Fusion}
\label{sec34}

Numerous studies, such as~\cite{wu2021cvt, yuan2023effective, zhang2022topformer, fan2023segtransconv}, have explored integrating features from both Transformers and CNNs. 
For example, SegTransConv~\cite{fan2023segtransconv} introduces a hybrid architecture combining Transformers and CNNs in series and parallel, yet it does not fully exploit the collaborative potential of both. 
Given the distinct characteristics and computational mechanisms of Transformers and CNNs, conventional element-wise addition or concatenation operations may not yield optimal results. 
A design leveraging the complementary strengths of both is therefore crucial for maximizing the representability of the extracted features and facilitating information recovery during decoding. 

In this paper, we introduce an effective strategy to bridge this gap. Our approach seamlessly combines the distinct types of features extracted by Transformers and CNNs 
through correlation-weighted integration. 
By fusing CNN and Transformer features with high correlation, we 
develop a new correlation-weighted Fusion (cwF) module. 


As depicted in Fig.~\ref{fig:cwf}, $T$ and $F$ denote intermediate features from the Transformer and CNNs, respectively. 
Initially, the Transformer feature $T$ is reshaped to match the same 
shape of the CNN feature $F$, which is followed by the post-concatenation operation of the two feature sets. 
To reduce the computational costs, depthwise separable convolution is employed for channel dimensional reduction. 
Subsequent to GAP and Sigmoid operations, a correlation 
coefficient matrix, denoted by $M$, is calculated. 
This matrix is then multiplied with the original features to obtain $\mathop F\limits^ \sim$ and $\mathop T\limits^ \sim$, which are added together to produce the final output $ Z $. 

This process can be 
expressed 
as 
\begin{equation}
    G = \text{Concat}[\text{Reshape}\left( {T|F} \right),F],
\end{equation}
where $G \in {\mathbb{R}^{ C_g  \times H_g \times W_g}}$ $\left(C_g = C_f + C_t\right)$), $Concat$ represents the concatenating operation, and $a|b$ represents the feature map of the size $a$ being restored to size $b$. Then, the correlation coefficient matrix $M$ can be computed 
as
\begin{equation}
    M = \delta \left( {{f_{1 \times 1}}\left( {\text{GAP}\left( {{f_{1 \times 1}}\left( {{f_{3 \times 3}}\left( G \right)} \right)} \right)} \right)} \right),
\end{equation}
where $M \in {\mathbb{R}^{C_m \times 1 \times 1}}$, $\delta$ is the Sigmoid function, $GAP$ represents global average pooling operation, and ${f_{{k_1} \times {k_2}}}$ denotes convolutional operation with a kernel size of ${k_1} \times {k_2}$. 

Thus, the resultant 
cwF features, denoted by ${Z_{\text{cwF}}}$, can be expressed as 
\begin{equation}
    {Z_{\text{cwF}}} = \varphi \left( {\mathop T\limits^ \sim   + \mathop F\limits^ \sim  } \right),\left\{ {\mathop T\limits^ \sim   = T \odot M,\mathop F\limits^ \sim   = F \odot M} \right\},
\end{equation}
where ${Z_{\text{cwF}}} \in {\mathbb{R}^{C_z \times H_z \times W_z}}$, $\varphi$ is the ReLU activation function, 
and $\odot$ represents element-wise multiplication.

It is noteworthy that feature correlation 
has also been explored in CTCNet~\cite{gao2023ctcnet}, where the correlation between the features derived from Transformers and CNNs is calculated. However, in CTCNet, the module merely concatenates the correlation with the Transformer and CNN features, which 
cannot 
effectively align these two types of 
features, potentially leading to 
performance degradation due to feature mismatch.

\begin{figure}[t]
 \centerline{\includegraphics[width=0.8\linewidth]{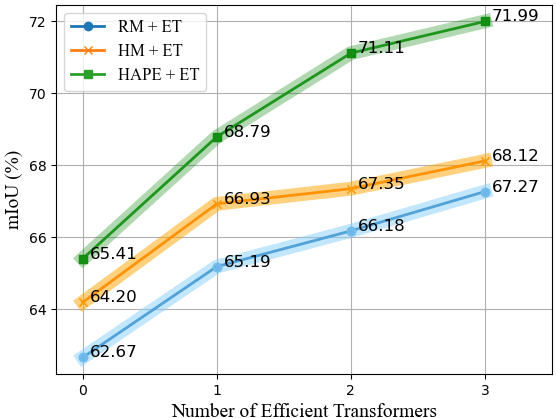}}
	\caption{Performance comparisons using different numbers of ETs with the baseline RM module, HM module, and our HAPE module obtained on the CamVid dataset.}
	\label{fig:num}
\end{figure}

\begin{table}[t]
\caption{Ablation studies on the HM and PEM components of the proposed HAPE module. 
\textit{Cit.}, \textit{Cam.}, and \textit{Param.} denote Cityscapes, CamVid, and Parameter, respectively.}
\begin{center}	
\setlength\tabcolsep{3pt}
\renewcommand{\arraystretch}{1.1}
\begin{tabular}{l|ccccc}
\toprule[0.35mm]
\multirow{2}{*}{\textbf{Architecture}} & \multirow{2}{*}{\textbf{Param. (K)}$\downarrow$} & \multicolumn{2}{c}{\textbf{FLOPs (G)}$\downarrow$}    & \multicolumn{2}{c}{\textbf{mIoU (\%)}$\uparrow$}           \\ \cline{3-4} \cline{5-6}
 &             & \multicolumn{1}{c}{Cit.} & Cam. & \multicolumn{1}{c}{Cit.} & Cam. \\ \hline \hline
Baseline (RM) & 424.912  & \multicolumn{1}{c}{10.166} & 5.957  &  
 \multicolumn{1}{c}{66.78}           & 62.67      \\ 

Baseline (HM) & 450.256  & \multicolumn{1}{c}{10.401} & 6.094  & \multicolumn{1}{c}{$68.25^{ \color{blue}{1.47\uparrow}}$}           & $64.20^{ \color{blue}{1.53\uparrow}}$        \\ 

Baseline (HAPE) & 482.512  & \multicolumn{1}{c}{10.402} & 6.095  & \multicolumn{1}{c}{$68.91^{ \color{blue}{2.13\uparrow}}$}           & $65.41^{ \color{blue}{2.74\uparrow}}$       \\ \hline \hline

\rowcolor{gray!20} HAFormer (Ours) & 602.298 & \multicolumn{1}{c}{11.051} & 6.475 &  \multicolumn{1}{c}{\textbf{$74.18^{ \color{blue}{7.40\uparrow}}$}}       & \textbf{$71.11^{ \color{blue}{8.44\uparrow}}$}        \\ 
\bottomrule[0.35mm]
\end{tabular}
\label{tab:hape}
\end{center}
\end{table}

\begin{table}[t]
\caption{Ablation studies on the impact of Dilation Rates.}
\begin{center}	
\setlength\tabcolsep{2pt}
\renewcommand{\arraystretch}{1.2}
\begin{tabular}{l|ccccl}
\toprule[0.35mm]
\multirow{2}{*}{\textbf{Architecture}}  & \multicolumn{4}{c}{\textbf{Dilation Rate}}                 & \multicolumn{1}{c}{\multirow{2}{*}{\textbf{mIoU (\%)}$\uparrow$}} \\ 
    & \multicolumn{1}{c}{Stage1}  & \multicolumn{1}{c}{Stage2}          
    & \multicolumn{1}{c}{Stage3}  & Stage4 & \multicolumn{1}{c}{}  \\ \hline\hline

\multirow{3}{*}{Baseline (HAPE)} & \multicolumn{1}{c}{(1,1,1)} & \multicolumn{1}{c}{(1,1,1,1,1,1)}   & \multicolumn{1}{c}{(1,1,1,1,1,1)}   & (1,1,1) 
& \multicolumn{1}{c}{68.91} \\ 

& \multicolumn{1}{c}{(2,2,2)} & \multicolumn{1}{c}{(2,2,2,2,2,2)}   
& \multicolumn{1}{c}{(2,2,2,2,2,2)}   & (2,2,2) & \multicolumn{1}{c}{$69.37^{ \color{blue}{0.46\uparrow}}$}   \\  

& \multicolumn{1}{c}{(2,2,2)} & \multicolumn{1}{c}{(4,4,8,8,16,16)} 
& \multicolumn{1}{c}{(4,4,8,8,16,16)} & (2,2,2)  & \multicolumn{1}{c}{$70.12^{ \color{blue}{1.21\uparrow}}$} \\ \hline

\multirow{3}{*}{HAFormer}     & \multicolumn{1}{c}{(1,1,1)} & \multicolumn{1}{c}{(1,1,1,1,1,1)}   & \multicolumn{1}{c}{(1,1,1,1,1,1)}  & (1,1,1) & \multicolumn{1}{c}{72.45} \\ 

& \multicolumn{1}{c}{(2,2,2)} & \multicolumn{1}{c}{(2,2,2,2,2,2)}   &\multicolumn{1}{c}{(2,2,2,2,2,2)}   & (2,2,2) & \multicolumn{1}{c}{$73.16^{ \color{blue}{0.71\uparrow}}$}\\

& \multicolumn{1}{c}{(2,2,2)} & \multicolumn{1}{c}{(4,4,8,8,16,16)} & \multicolumn{1}{c}{(4,4,8,8,16,16)} & (2,2,2)& \multicolumn{1}{c}{\cellcolor{gray!30}\textbf{$74.18^{ \color{blue}{1.73\uparrow}}$}} \\ \hline \hline

\rowcolor{gray!20} HAFormer (ours) & \multicolumn{1}{c}{(2,2,2)}  & \multicolumn{1}{c}{(4,4,8,8,16,16)} & \multicolumn{1}{c}{(4,4,8,8,16,16)}     & \multicolumn{1}{c}{(2,2,2)} & \multicolumn{1}{c}{\cellcolor{gray!30}\textbf{74.18}} \\ 
\bottomrule[0.35mm]
\end{tabular}
\vspace{-1em}
\label{tab:dr}
\end{center}
\end{table}

\begin{table}[t]
\caption{Performance comparison between using TT and ET in HAPE on the Cityscapes ($512 \times 1024$) and CamVid ($360 \times 480$) datasets.}
\begin{center}
\setlength\tabcolsep{1.4pt}
\renewcommand{\arraystretch}{1.1}
\begin{tabular}{c|ccccccccc}
\toprule[0.35mm]
\multirow{2}{*}{Architecture} & \multirow{2}{*}{TT} & \multirow{2}{*}{ET} & \multirow{2}{*}{Param. (K)$\downarrow$} & \multicolumn{2}{c}{FLOPs (G)}$\downarrow$ & \multicolumn{2}{c}{Speed (FPS)}$\uparrow$ & \multicolumn{2}{c}{mIoU (\%)}$\uparrow$ \\ \cline{5-6} \cline{7-8}\cline{9-10}
&                     &                     &                  & Cit.          & Cam.         & Cit.       & Cam.    & Cit.       & Cam.      \\ \hline\hline
\multirow{2}{*}{HAFormer}     
& \checkmark   &  & 760.293   & 13.341     &  8.135    & 56    & 77  & 74.66   & 71.47     \\
&   &  \checkmark    &  602.298  & 11.051  & 6.095   & 105   & 118  & 74.18     & 71.11
\\ 
\bottomrule[0.35mm]
\end{tabular}
\label{tab:Trans}
\end{center}
\end{table}

\section{Experiments}
\label{sec4}

To demonstrate the effectiveness of our HAFormer and its individual modules qualitatively and quantitatively, comparative experiments are conducted on benchmark datasets and compared with state-of-the-art (SOTA) approaches. 
In this section, we first 
outline the datasets, loss functions, hardware platform configuration, and parameter settings used in our experiments. 
Then, 
we present the series of ablation experiments conducted to validate the effectiveness of the individual modules. 
Finally, 
comparative experiments are conducted to demonstrate the superiority of our approach over 
the SOTA approaches. 

\subsection{Datasets}
\label{sec41}

Our HAFormer model is designed to tackle challenges related to scale variations and contextual information in street scenes. 
The Cityscapes~\cite{cordts2016cityscapes} and CamVid~\cite{brostow2008segmentation} datasets are two prominent benchmarks widely utilized in street scene segmentation research. 
Hence, to showcase the efficacy of our model, we conducted a series of comprehensive empirical evaluations 
on these two datasets.

\textbf{Cityscapes.} This dataset comprises 5,000 high-quality images annotated at the pixel level. Captured from various urban settings in 50 cities, these images have a resolution of $2,048 \times 1,024$ and primarily depict driving scenes. The dataset is divided 
into three subsets: 2,975 images for training, 500 for validation, and 1,525 for testing. 
While the dataset includes labels for 34 categories, our study focuses specifically on 19 essential semantic categories. We utilize the Cityscapes' built-in tools to adjust the labels to suit our research needs.

\textbf{CamVid.} This is a public dataset of urban road scenes released by the University of Cambridge. The images, with a resolution of $960 \times 720$, are captured from a driving perspective, increasing the diversity of observed targets. With over 700 labeled images, the dataset is suitable 
for supervised learning. The CamVid dataset usually employs 11 common categories for evaluating segmentation accuracy. 
These categories offer a thorough representation of objects in urban road scenes, making them 
a valuable resource for 
research. 

\subsection{Implementation Details}
\label{sec42}

The HAFormer model is executed on a single RTX 2080 Ti GPU card with 12GB memory, using CUDA 10.1 and PyTorch 1.8.1. The architecture is trained from scratch without any pre-trained models.  
We employ Stochastic Gradient Descent (SGD) with a momentum of 0.9 and a weight decay of $1e - 5$, 
along with the ``Poly” learning rate policy for optimization. 

For Cityscapes, the initial learning rate is $4.5e - 2$, and the batch size is set to 5 to maximize GPU memory usage. For CamVid, the initial learning rate is $1e - 3$, with a batch size of 8. 
Following the existing practice, 
we apply 
data augmentation techniques including 
horizontal flipping, random scaling, and random cropping to introduce diversity in the training data, with random scales ranging from 0.25 to 2.0 and the cropping size of $512 \times 1024$ for Cityscapes over 1,000 epochs. No post-processing 
is applied for a fair comparison.


Finally, following the existing practice, the performance is quantitatively evaluated using the averaged mean Intersection-over-Union (mIoU) across all categories, 
as well as the parameter counts, FLOPs and GPU usage, and processing speed. 

\begin{figure*}[t]
	\centerline{\includegraphics[width=18cm]{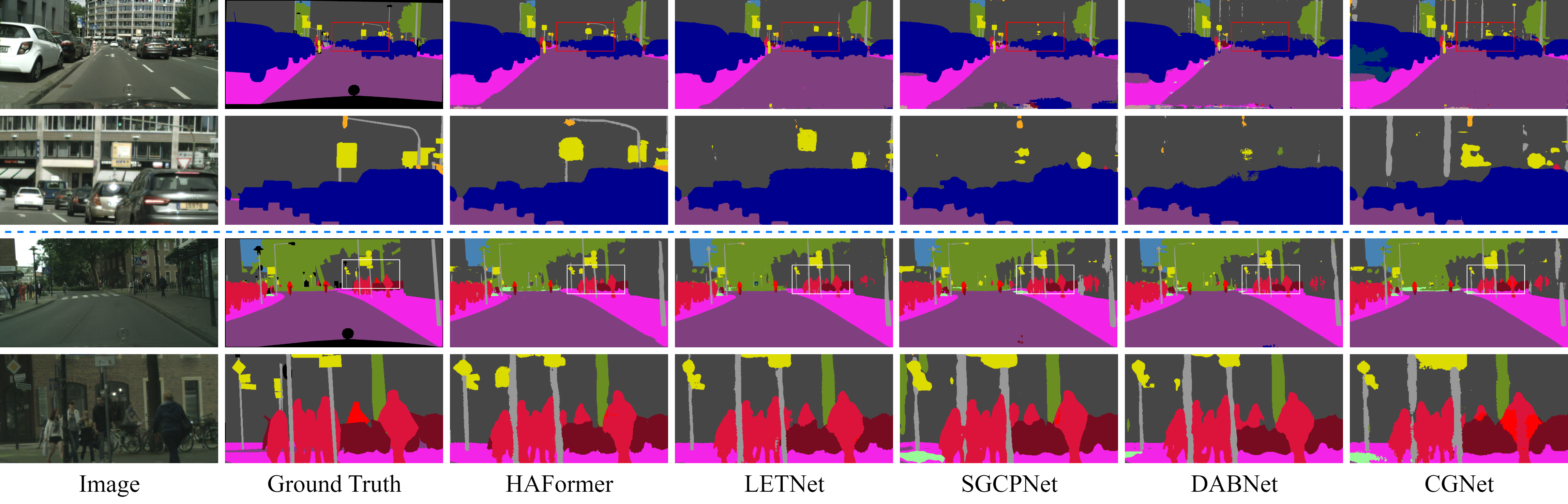}}
	\caption{Visual results on Cityscapes dataset. From left to right: original images, ground truths, predictions of \textbf{HAFormer}, LETNet~\cite{xu2023lightweight}, SGCPNet~\cite{hao2022real}, DABNet~\cite{li2019dabnet}, CGNet~\cite{wu2020cgnet}. 
    Note that two examples are shown. In each group, the first row visualizes the overall segmentation results, while the second row visualizes the zoom-in of the small areas enclosed in rectangles. 
    }
    \vspace{-1em}
	\label{fig:city}
\end{figure*}

\subsection{Ablation Studies}
\label{sec43}

In this part, we conduct a series of ablation experiments to validate the effectiveness of each module in our method. 

\textbf{Ablation Study of the HAPE Module.}
%

In our HAPE module (see Section~\ref{sec32}), we proposed four parallel convolution operations to capture image features across various hierarchies comprehensively. 
This is then followed by the PEM, designed to enhance the feature representability through a content-aware spatial-attention mechanism. 
In this section, we show the effectiveness of the hierarchical approach (denoted as ``HM'') and the PEM approach of our HAPE module, respectively.

The 
baseline model used for comparison is structured as a single-line type (as shown in Fig.~\ref{fig:base}), incorporating the standard Residual Modules (RMs). 
To showcase the performance gains brought by the HM and PEM, we first substitute the RM of the baseline model with the HM module, 
omitting the PEM part, 
and then include both the HM and PEM modules to test the effectiveness of the entire HAPE module.

Table~\ref{tab:hape} highlights the superior performance of the HM, showcasing mIoU gains of 1.47\% and 1.53\% over the RM. The HM excels in extracting robust features, facilitating deep semantic information extraction effectively. 
Moreover, the multi-scale structure significantly enhances the model's performance in feature extraction and small object recognition. 
Introducing the PEM 
further enhances segmentation accuracy by 2.13\% and 2.74\% on both datasets. 

Throughout this experiment, the dilated convolution rates are set to 1 in both HM and HAPE to ensure a fair comparison. 
Fig.~\ref{fig:num} also verifies 
the efficacy of our HAPE module when being integrated with the Transformer module.

\textbf{Ablation Study of the Dilation Rates.}
In this section, we explore how the chosen dilation rates impact segmentation performance. With a consistent number of modules, a larger 
dilation rate expands 
the receptive field, 
allowing the model to perceive a broader scope, 
and hence is essential for comprehensive feature extraction. 

Results shown in Table~\ref{tab:dr} reveal that transitioning the dilation rate from all 1s to all 2s (the first two rows) in dilated convolution boosts mIoU by about 0.5\%. 
Further, by progressively increasing the dilated convolution rate in Stages 2 and 3, we observe performance enhancements of 1.21\% and 1.73\% on the two datasets. 
Hence, to preserve spatial details, in our approach, we allocate three modules in Stages 1 and 4 while employing six modules in Stages 2 and 3 to capture intricate semantic information within the network's depth. This strategy optimizes calculations for the transformer encoder, improving long-range dependency modeling.

\begin{table}[t]
\caption{Ablation studies of the ET, the number of ETs in HAPE (``L''), and the cwF module on the Cityscapes dataset.}
\begin{center}
\setlength\tabcolsep{1.5pt}
\renewcommand{\arraystretch}{1.1}
\begin{tabular}{l|cccccc}
\toprule[0.35mm]
\multirow{2}{*}{\textbf{Architecture}}   & \multicolumn{3}{c}{\textbf{Fusion}}  & \multirow{2}{*}{\textbf{Param. (K)}$\downarrow$} & \multirow{2}{*}{\textbf{FLOPs (G)}$\downarrow$} & \multirow{2}{*}{\textbf{mIoU (\%)}$\uparrow$} \\ 
    & \multicolumn{1}{c}{Add} & \multicolumn{1}{c}{Concat} & \textbf{cwF} & && \\ \hline\hline \rule{0pt}{8pt}
    
Baseline (HAPE) & \multicolumn{1}{c}{$-$}    & \multicolumn{1}{c}{$-$}   & $-$    
&482.512 & 10.402  & 70.12 \\ \hline \rule{0pt}{8pt} 

\multirow{3}{*}{HAFormer (L=1)} & \multicolumn{1}{c}{$\checkmark$}    & \multicolumn{1}{c}{}       &     
&549.364    &10.780    & $71.23^{ \color{blue}{1.11\uparrow}}$   \\  \rule{0pt}{8pt} 

& \multicolumn{1}{c}{}    & \multicolumn{1}{c}{$\checkmark$}       &     
&585.556    &12.648    & $71.66^{ \color{blue}{1.54\uparrow}}$   \\  \rule{0pt}{8pt} 

& \multicolumn{1}{c}{}    & \multicolumn{1}{c}{}       &$\checkmark$     
&554.742    &10.952    & $72.50^{ \color{blue}{2.38\uparrow}}$    \\ \hline \rule{0pt}{8pt}
    
\multirow{3}{*}{HAFormer (L=2)} & \multicolumn{1}{c}{$\checkmark$}    & \multicolumn{1}{c}{}       &     
& 596.920  &10.879    & $72.28^{ \color{blue}{2.16\uparrow}}$ \\ \rule{0pt}{8pt}

& \multicolumn{1}{c}{}    & \multicolumn{1}{c}{$\checkmark$}       &     
& 633.112  &12.747  & $73.17^{ \color{blue}{3.05\uparrow}}$     \\ \rule{0pt}{8pt}

& \multicolumn{1}{c}{}    & \multicolumn{1}{c}{}       & $\checkmark$    
&602.298   &11.051  & \cellcolor{gray!30}\textbf{$74.18^{ \color{blue}{4.06\uparrow}}$}       \\
\hline

\rowcolor{gray!20}HAFormer (ours) &  \multicolumn{1}{c}{}   &  \multicolumn{1}{c}{}  
& $\checkmark$    
& 602.298  & 11.051  &  \cellcolor{gray!30}{\textbf{74.18}}      \\ 
\bottomrule[0.35mm]
\end{tabular}
\vspace{-1em}
\label{tab:ET}
\end{center}
\end{table}

\begin{table*}[t]
\caption{Comparisons on the Cityscapes test dataset. ``$*$" indicates that the method utilizes multiple graphics cards. ``$-$" means that the result is not provided in the corresponding methodology.}
\begin{center}
\setlength\tabcolsep{6pt}
\renewcommand{\arraystretch}{1.1}
\scalebox{0.9}{
\begin{tabular}{c|l|lccccccc}
\toprule[0.35mm]
& \textbf{Methods} 
& \textbf{Year} 
& \textbf{Resolution} 
& \textbf{Backbone} 
& \textbf{Param. (M) $\downarrow$ } & \textbf{FLOPs (G) $\downarrow$} 
& \textbf{GPU} 
& \textbf{Speed (FPS) $\uparrow$} 
& \textbf{mIoU (\%) $\uparrow$} \\ \hline \hline
\multirow{10}{*}{\rotatebox{90}{\textbf{Large Model}}}  

& SegNet~\cite{badrinarayanan2017segnet}  &2017 TPAMI   & $360 \times 640$  &  VGG-16   & 29.50 & 286.0  & $*$    &  17.0   & 57.0 \\
& DeepLab~\cite{chen2017deeplab}  &2017 TPAMI  & $512 \times 1024$ & ResNet 101 & 262.10 & 457.8  & $*$    & 0.3  & 63.5  \\ 
& PSPNet~\cite{zhao2017pyramid}   &2017 CVPR  & $713 \times 713$  & ResNet-101  & 68.10    & 2048.9 & $*$   & 1.2  &  78.5         \\  
& DeepLab-V3+~\cite{chen2018encoder}  & 2018 ECCV & $-$  &  MobileNet-V2 &62.70 & 2032.3  & $*$    & 1.2   &  80.9         \\  
& DANet~\cite{fu2019dual}  & 2019 CVPR   & $1024 \times 1024$  & ResNet-101 & 66.60   & 1111.8 & $*$     & 4.0   & 81.5           \\ 
& SETR~\cite{zheng2021rethinking}   & 2021 CVPR   & $768 \times 768$    & ViT-Large  & 318.30  & $-$ & $*$    & 0.5 &  82.2         \\ 
& HRViT~\cite{gu2022multi}  & 2022 CVPR & $1024 \times 2048$ & HRViT     & 28.60 & 66.8 & $*$   & $-$  & 83.2           \\ 
& SegFormer~\cite{xie2021segformer} &2021 NIPS & $1024 \times 2048$ & MiT-B5     & 84.70  & 1447.6 & $*$    & 2.5  & 84.0           \\ 
& Lawin~\cite{yan2022lawin} &2022 arxiv & $1024 \times 1024$ & Swin-L     & $-$  & 1797.0 & $*$    & $-$  & \cellcolor{gray!30}\textbf{84.4}           \\ 
& DDPS-SF~\cite{lai2023denoising} & 2023 arxiv  & $1024 \times 2048$ & MiT-B5 & 122.80 &  $-$   & $*$   & $-$    & 82.4          \\ \hline \hline 
\multirow{20}{*}{\rotatebox{90}{\textbf{Lightweight Model}}} 

& ENet~\cite{paszke2016enet}    & 2016 arxiv  & $512 \times 1024$ &  No        & \cellcolor{gray!30}\textbf{0.36} & 3.8 & Titan X    & 135 & 58.3          \\  
& ESPNet~\cite{mehta2018espnet}   &2018 ECCV  & $512 \times 1024$  & ESPNet   & \cellcolor{gray!30}\textbf{0.36}  & $-$   & Titan XP  & 113 & 60.3         \\ 
& NDNet~\cite{yang2020ndnet}   &2021 TITS   & $512 \times 1024$    & No         & 0.50   & \cellcolor{gray!30}\textbf{3.5}   & Titan X    & 101  &  61.1         \\ 
& CGNet~\cite{wu2020cgnet}   &2021 TIP   & $360 \times 640$    & No         & 0.50   & 6.0   & V100    & 120  &  64.8         \\ 
& ERFNet~\cite{romera2017erfnet}  & 2017 TITS  & $512 \times 1024$  & No        & 2.10  & $-$   & Titan X    & 42  & 68.0          \\
& ICNet~\cite{zhao2018icnet}   & 2018 ECCV  & $1024 \times 2048$ & PSPNet-50  &26.50 & 28.3  & $-$    & 30   & 69.5   \\ 
& DABNet~\cite{li2019dabnet}   & 2019 BMVC  & $1024 \times 2048$ & No        & 0.76  & 42.4  & 1080 Ti     & 28  & 70.1          \\    
& FPENet~\cite{liu2019feature}   &2019 BMVC & $512 \times 1024$   & No         & 0.40  & 12.8  & Titan V    & 55 & 70.1          \\  
& LEDNet~\cite{wang2019lednet}  &2019 ICIP   & $512 \times 1024$   & No         & 0.94  & $-$   & 1080 Ti  & 71  & 70.6          \\ 
& FBSNet~\cite{gao2022fbsnet}  & 2023 TMM  & $512 \times 1024$  &  No        & 0.62  & 9.7    & 2080 Ti  & 90   & 70.9 \\ 
& SGCPNet~\cite{hao2022real}  & 2022 TNNLS & $1024 \times 2048$  &  MobileNet       & 0.61   & 4.5   & 1080 Ti   & 103  &70.9     \\ 
& MSCFNet~\cite{gao2021mscfnet}  & 2022 TITS & $512 \times 1024$  & No         & 1.15   & 17.1    & Titan XP  & 50   & 71.9   \\  
& SegFormer~\cite{xie2021segformer} &2021 NIPS & $512 \times 1024$ & MiT-B0      & 3.80     & 17.7  & V100 & 48  & 71.9   \\
& MLFNet~\cite{fan2022mlfnet} &2023 TIV & $512 \times 1024$ & ResNet-34      & 13.03     & 15.5  & Titan XP & 72  & 72.1   \\ 
& BiseNet-V2~\cite{yu2021bisenet}&2021 IJCV & $1024 \times 2048$ & Xception     & 3.40     & 21.2  & 1080 Ti & \cellcolor{gray!30}\textbf{156}  & 72.6   \\
& PCNet~\cite{lv2021parallel} &2022 TITS & $1024 \times 2048$ & Scratch      & 1.63     & 11.8  & 2080 Ti & 72  & 72.7   \\ 
& MGSeg~\cite{he2021mgseg} &2021 TIP & $1024 \times 1024$ & ShuffleNet-V2      & 4.50     & 16.2  & Titan XP & 101  & 72.7   \\ 
& LETNet~\cite{xu2023lightweight}  & 2023 TITS & $512 \times 1024$   &  No        & 0.95    & 13.6    & 3090    & 150   &72.8           \\
& SegTransConv~\cite{fan2023segtransconv}  & 2023 TITS & $512 \times 1024$   &  STDC        & 7.00    & 10.2    & 3090    & 57   &73.0           \\
& PMSDSEN~\cite{liu2023efficient}  & 2023 ACM MM& $512 \times 1024$   &  No        & 0.92   & 10.2    & $-$    & 53   &73.2           \\
& EFRNet-16~\cite{li2022efrnet}  &2022 TMM & $512 \times 1024$ & EAA & 1.44 & 25.1 & Titan X & 58 &\cellcolor{gray!30}\textbf{74.3} \\ 
\cline{2-9}
& \cellcolor{gray!20}\textbf{HAFormer (Ours)}
&\cellcolor{gray!20}-  
& \cellcolor{gray!20}$512 \times 1024$ 
&  \cellcolor{gray!20}No         & \cellcolor{gray!20}0.60    &  \cellcolor{gray!20}11.1   & \cellcolor{gray!20}2080 Ti    &  \cellcolor{gray!20}105 
& \cellcolor{gray!30}\textbf{74.2}           \\ 
\bottomrule[0.35mm]
\end{tabular}
}
\label{tab:city}
\end{center}
\end{table*}

\begin{table*}[t]
\caption{Comparisons with other methods about per-class results on the Cityscapes test set. Roa: Road, Sid: Sidewalk, Bui: Building, Wal: Wall, Fen: Fence, Pol: Pole, TLi: Traffic Light, TSi: Traffic Sign, Veg: Vegtation, Ter: Terrain, Sky: Sky, Ped: Pedestrain, Rid: Rider, Car: Car, Tru: Truck, Mot: Motorcycle, Bic: Bicycle.}

\begin{center}
\setlength{\tabcolsep}{3.8pt}
\renewcommand{\arraystretch}{1.1}
\vspace{-1em}
\begin{tabular}{l|ccccccccccccccccccc|c}
\toprule[0.35mm]
\textbf{Methods} & \textbf{Roa} & \textbf{Sid} & \textbf{Bui} & \textbf{Wal} & \textbf{Fen} & \textbf{Pol} & \textbf{TLi} & \textbf{TSi} & \textbf{Veg} & \textbf{Ter} & \textbf{Sky} & \textbf{Ped} & \textbf{Rid} & \textbf{Car} & \textbf{Tru} & \textbf{Bus} & \textbf{Tra} & \textbf{Mot} & \textbf{Bic} & \textbf{mIoU} \\ \hline \hline

SegNet~\cite{badrinarayanan2017segnet} & 96.4&73.2 &84.0 &28.4 &29.0 &35.1 &39.8 &45.1 &87.0 &63.8 &91.8 &62.8 &42.8 &89.3 &38.1 &43.1 &44.1 &35.8 &51.9 &57.0  \\ 

ENet~\cite{paszke2016enet} & 96.3&74.2 &75.0 &32.2 &33.2 &43.4 &34.1 &44.0 &88.6 &61.4 &90.6 &65.5 &38.4 &90.6 &36.9 &50.5 &48.1 &38.8 &55.4 &58.3  \\ 

ESPNet~\cite{mehta2018espnet} & 97.0&77.5 &76.2 &35.0 &36.1 &45.0 &35.6 &46.3 &90.8 &63.2 &92.6 &67.0 &40.9 &92.3 &38.1 &52.5 &50.1 &41.8 &57.2 &60.3  \\ 

CGNet~\cite{wu2020cgnet} & 95.5 &78.7 &88.1 &40.0 &43.0 &54.1 &59.8 &63.9 &89.6 &67.6 &92.9 &74.9 &54.9 &90.2 &44.1 &59.5 &25.2 &47.3 &60.2 &64.8 \\ 

ERFNet~\cite{romera2017erfnet} & 97.7 &81.0 &89.8 &42.5 &48.0 &56.3 &59.8 &65.3 &91.4 &68.2 &94.2 &76.8 &57.1 &92.8 &50.8 &60.1 &51.8 &47.3 &61.7 &68.0 \\ 

LEDNet~\cite{wang2019lednet} & 98.1 &79.5 &91.6 &47.7 &49.9 &62.8 &61.3 &72.8 &92.6 &61.2 &\cellcolor{gray!30}\textbf{94.9} &76.2 &53.7 &90.9 &64.4 &64.0 &52.7 &44.4 &71.6 &70.6  \\ 

FBSNet~\cite{gao2022fbsnet} & 98.0 &83.2 &91.5 &50.9 &53.5 &62.5 &\cellcolor{gray!30}\textbf{67.6} &71.5 &\cellcolor{gray!30}\textbf{92.7} &\cellcolor{gray!30}\textbf{70.5} &94.4 &82.5 &\cellcolor{gray!30}\textbf{63.8} &93.9 &50.5 &56.0 &37.6 &56.2 &70.1 &70.9  \\ 

LARNet~\cite{hu2023lightweight} & 98.0 &82.2 &90.7 &48.9 &44.7 &57.2 &62.8 &67.2 &92.0 &68.6 &94.7 &79.3 &59.8 &93.9 &54.4 &73.9 &\cellcolor{gray!30}\textbf{61.3} &54.0 &66.1 &71.1  \\ 

MSCFNet~\cite{gao2021mscfnet} & 97.7 &82.8 &91.0 &49.0 &52.5 &61.2 &67.1 &71.4 &92.3 
&70.2 &94.3 &\cellcolor{gray!30}\textbf{82.7} &62.7 &94.1 &50.9 &66.1 &51.9 &\cellcolor{gray!30}\textbf{57.6} &70.2 &71.9  \\ 

LETNet~\cite{xu2023lightweight} & 98.2 &\cellcolor{gray!30}\textbf{83.6} &91.6 &50.9 &53.7 &61.0 &66.7 &70.5 &92.5 &70.5 &\cellcolor{gray!30}\textbf{94.9} &82.3 &61.7 &\cellcolor{gray!30}\textbf{94.4} &55.0 &72.4 &57.0 &56.1 &69.3 &72.8  \\ \hline\hline

\rowcolor{gray!20} \textbf{HAFormer (Ours)} & \cellcolor{gray!30}\textbf{98.4} & 82.6 &\cellcolor{gray!30}\textbf{91.7} &\cellcolor{gray!30}\textbf{57.2} &\cellcolor{gray!30}\textbf{61.1} &\cellcolor{gray!30}\textbf{63.0} &62.2 &\cellcolor{gray!30}\textbf{74.3} &91.8 &61.7 &93.8 &79.3 &56.4 &93.7 &\cellcolor{gray!30}\textbf{66.9} &\cellcolor{gray!30}\textbf{80.1} &65.3 &56.7 &\cellcolor{gray!30}\textbf{73.5} & \cellcolor{gray!30}\textbf{74.2}  \\ 
\bottomrule[0.35mm]
\end{tabular}
\vspace{-1em}
\label{tab:perclass}
\end{center}
\end{table*}

\textbf{Ablation Study of the Efficient Transformer.} 

As detailed in Section~\ref{sec33}, another key contribution we made in the HAFormer is the Efficient Transformer (ET) module, which reduces the dimension of features by projecting them into an optimal latent embedding space before calculating self-attention. 
Table~\ref{tab:Trans} showcases the performance gains brought by the ET module over the traditional Transformer (denoted as ``TT'') in terms of segmentation accuracy and computation complexity on Cityscapes and CamVid datasets. 

As shown in Table~\ref{tab:Trans}, the ET design demonstrates a superior balance between efficiency and accuracy. Compared to the traditional Transformer ``TT'', ET 
achieves an 18\% reduction in parameter count and a 17\% decrease in computational load, with only a slight mIoU loss of 0.4\%. This results in a more efficient model with minimal impact on performance, and it even offers faster inference speed. 
In addition, 
the results in Table~\ref{tab:ET} also reveal the 
significant enhancement upon integrating features learned through the Transformer, with a remarkable 2.16\% boost in mIoU. 
This underscores the Transformer's exceptional ability to capture long-range dependencies, a feature that the CNN alone cannot achieve.

Additionally, in the proposed HAFormer, the number of ET layers $L$ is deliberately limited to 2, considering computation hardware constraints and also aiming to achieve 
the best balance under constraints. 
Although stacking more ET layers could 
yield better accuracy results, as shown in Fig.~\ref{fig:num}, 
the performance gains slow down dramatically when $L$ is greater than 2. 
Moreover, adding excessive ET layers on a high-resolution dataset like Cityscapes may negatively impact parameters, computations, and inference speed, potentially causing overfitting.

\textbf{Ablation Study of the Correlation-weighted Fusion.} 
To address the feature mismatch issue between CNNs and Transformers 
and ensure effective feature restoration during decoding, in Section~\ref{sec34} we introduced the cwF mechanism. 
Table~\ref{tab:ET} compares the results obtained with our cwF method and two other fusion techniques, \textit{i.e.}, element-wise addition, and concatenation. 
The table illustrates enhanced segmentation accuracy when integrating CNN and Transformer features using all three fusion methods. Notably, our cwF achieves a performance improvement of 2.38\% over the baseline with one ET layer and 4.06\% gain with two layers stacked.

Moreover, from Table~\ref{tab:ET} we observe that (a) Compared to the simple element-wise addition fusion scheme, our cwF shows performance gains of 1.27\% and 1.90\% in the two cases with only a slight increase in parameter count and FLOPs; (b) Our cwF presents mIoU gains of 0.84\% and 1.01\% over the computationally expensive concatenation operation, respectively, while achieving about 5\% reduction in parameter count and 15\% decrease in computational load. These experimental outcomes further demonstrate the effectiveness of our cwF.

\subsection{Comparisons with SOTA Methods} 
\label{sec44}

In this section, we extensively assess and compare the performance and efficiency of our method against some state-of-the-art approaches to showcase the advantages of our proposed method. Our evaluation centers on three key aspects: segmentation accuracy, model parameters, and floating-point operations (FLOPs). 

\begin{figure*}[t]
	\centerline{\includegraphics[width=18cm]{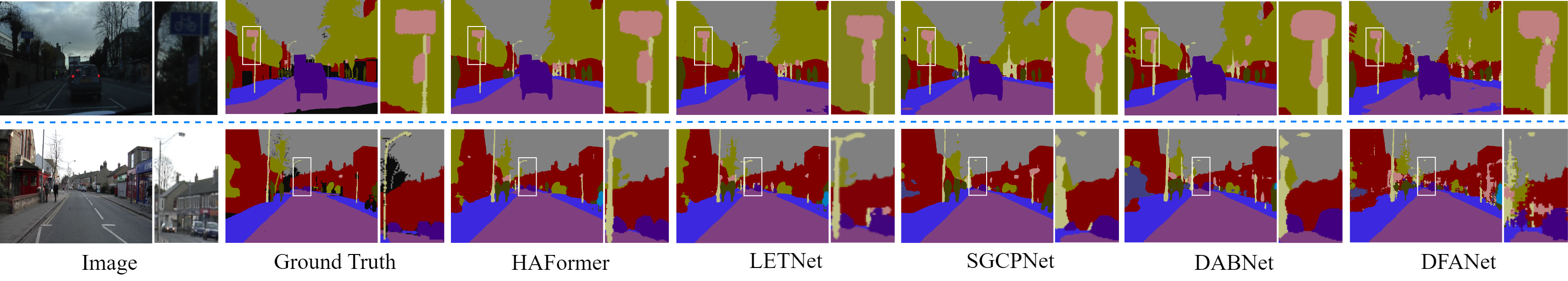}}
	\caption{Visual results obtained on the CamVid dataset. From left to right: original images, ground truths, predictions obtained with \textbf{HAFormer}, LETNet~\cite{xu2023lightweight}, SGCPNet~\cite{hao2022real}, DABNet~\cite{li2019dabnet}, DFANet~\cite{li2019dfanet}. Next to each prediction result is a partially enlarged detail map.}
	\label{fig:camd}
\end{figure*}

\begin{table*}[!t]
\caption{Comparisons on the CamVid test dataset.}
\begin{center}
\setlength\tabcolsep{10 pt}
\renewcommand{\arraystretch}{1.1}
\vspace{-1em}
\begin{tabular}{l|cccccc}
\toprule[0.35mm]
\textbf{Methods}  & \textbf{Resolution} & \textbf{Backbone} & \textbf{Param. (M) $\downarrow$} & \textbf{GPU} & \textbf{Speed (FPS) $\uparrow$} & \textbf{mIoU (\%) $\uparrow$} \\ \hline \hline
ENet~\cite{paszke2016enet}     & $360 \times 480$   & No        & \cellcolor{gray!30}\textbf{0.36}  & Titan X  & 98 & 51.3 \\ 
SegNet~\cite{badrinarayanan2017segnet}   & $360 \times 480$   & VGG-16    & 29.50 & Titan X  & 60     & 55.6      \\ 
NDNet~\cite{yang2020ndnet}    & $360 \times 480$   & No        & 0.50  & Titan X  & 78     & 57.2      \\ 
DFANet~\cite{li2019dfanet}   & $720 \times 960$   & Xception  & 7.80  & Titan X  & 120    & 64.7      \\ 
DABNet~\cite{li2019dabnet}   & $360 \times 480$   & No        & 0.76  & 1080 Ti   & 136    & 66.4      \\ 
ICNet~\cite{zhao2018icnet}    & $720 \times 960$   & PSPNet-50 & 7.80  & Titan X  & 28     & 67.1      \\ 
LARNet~\cite{hu2023lightweight}   & $360 \times 480$   & No        & 0.95  & 2080 Ti  & 204    & 67.1      \\ 
EFRNet-16~\cite{li2022efrnet}   & $720 \times 960$   & EAA        & 1.44  & Titan X   & 154    & 68.2      \\ 
FBSNet~\cite{gao2022fbsnet}   & $360 \times 480$   & No        & 0.62  & 2080 Ti  & 120    & 68.9      \\ 
SGCPNet~\cite{hao2022real}  & $720 \times 960$   & No        & 0.61  & 1080 Ti   & \cellcolor{gray!30}\textbf{278} & 69.0 \\ 
MLFNet~\cite{fan2022mlfnet}  & $720 \times 960$   & ResNet-34        & 13.03  & Titan XP   & 57 & 69.0 \\ 
MSCFNet~\cite{gao2021mscfnet}  & $360 \times 480$   & No        & 1.15  & Titan XP & 110    & 69.3      \\ 
AGLNet~\cite{zhou2020aglnet}   & $360 \times 480$   & No        & 1.12  & 1080 Ti   & 90     & 69.4      \\ 
LETNet~\cite{xu2023lightweight}   & $360 \times 480$   & No        & 0.95  & 3090     & 200    & 70.5      \\
MGSeg~\cite{he2021mgseg}    & $736 \times 736$   & ResNet-18 & 13.3   & Titan XP & 127    & \textbf{72.7}      \\

\hline\hline
\multirow{2}{*}{ \textbf{HAFormer (Ours)}} & $360 \times 480$   & No     & 0.60  & 2080 Ti  
& 118    & \cellcolor{gray!30}\textbf{71.1} \\
& $720 \times 960$   & No     & 0.60  & 2080 Ti  
& 83    & \cellcolor{gray!30}\textbf{71.9}
\\ \bottomrule[0.35mm]
\end{tabular}
\vspace{-1em}
\label{tab:camd}
\end{center}
\end{table*}

\begin{table}[!t]
\caption{Comparisons of run-time and inference speed of the proposed HAFormer with other approaches.}
\begin{center}
\setlength\tabcolsep{2.3pt}
\renewcommand{\arraystretch}{1.1}
\vspace{-1em}
\begin{tabular}{l|c|cccc}
\toprule[0.35mm]
\multirow{3}{*}{Methods} & \multirow{3}{*}{Param. (M)$\downarrow$}
& \multicolumn{4}{c}{RTX 2080Ti}        \\ \cline{3-6}
&  & \multicolumn{4}{c}{$512 \times 1024$} \\
&   & FLOPs (G) $\downarrow$   & ms $\downarrow$   & fps $\uparrow$     & mIoU (\%) $\uparrow$      \\ \hline \hline
ENet~\cite{paszke2016enet}  & \cellcolor{gray!30}\textbf{0.36}& 3.8    & 13 & 112  &  58.3      \\
NDNet~\cite{yang2020ndnet}    & 0.50    & \cellcolor{gray!30}\textbf{3.5}     &  8     & 119     & 61.1        \\
DABNet~\cite{li2019dabnet} & 0.76   & 10.5      &    \cellcolor{gray!30}\textbf{7}    & \cellcolor{gray!30}\textbf{139}  & 69.3   \\ \hline 
FBSNet~\cite{gao2022fbsnet}  & 0.62  & 9.7     &    11       &  90   & 70.9        \\
MSCFNet~\cite{gao2021mscfnet}   & 1.15   & 17.1    &  18   &  56  & 71.9   \\
MLFNet~\cite{fan2022mlfnet}  & 13.03  & 15.5   &  14  &   72     & 72.2        \\
LETNet~\cite{xu2023lightweight}  & 0.95  & 13.6     &  11    & 91     & 72.8  \\ 
SegTransConv~\cite{fan2023segtransconv} & 7.00   &  10.2    & 17   & 58    & 73.0  \\
\hline\hline        
\rowcolor{gray!20}\textbf{HAFormer (ours)} & 0.60  & 11.1  & 10  & 105 & \textbf{74.2}\\
\bottomrule[0.35mm]
\end{tabular}
\vspace{-1em}
\label{tab:efficiency}
\end{center}
\end{table}

\textbf{Evaluation Results on Cityscapes.} Quantitative comparisons with advanced semantic segmentation methods on the Cityscapes test set are presented in Table~\ref{tab:city}. Per-class results are detailed in Table~\ref{tab:perclass}, and visualization outcomes are displayed in Fig.~\ref{fig:city}. To ensure fairness, no augmentation techniques are used during testing, and data for other networks are referenced from pertinent sources. Contemporary semantic segmentation models fall into two main categories: those emphasizing larger size and higher precision, and those prioritizing real-time practicality with a balance between accuracy and efficiency. 

While larger models achieve high accuracy, their FLOPs and speed lag behind lightweight models, making them unsuitable for real-time processing on devices with limited resources. In contrast, lightweight models like ENet~\cite{paszke2016enet}, ESPNet~\cite{mehta2018espnet}, CGNet~\cite{wu2020cgnet}, and FPENet~\cite{liu2019feature} are computationally efficient. Despite their reduced parameter count, their overall performance, especially in accuracy, is lacking. In terms of accuracy, EFRNet-16~\cite{li2022efrnet} shows similarities to our results. However, it is noteworthy that its parameter count and GFlops are 2 times greater than ours. Apparently, our model requires fewer parameters and computations, highlighting the efficiency of our approach.

\textbf{Evaluation Results on CamVid.} To further validate the effectiveness and generalization capacity of our model, we compared it with other lightweight methods on the CamVid dataset, as shown in Table~\ref{tab:camd}. While MGSeg~\cite{he2021mgseg} excels in accuracy, surpassing our method by 1.6 points, it does so at the cost of having 22 times more parameters than ours, indicating an unfavorable trade-off. On the other hand, SGCPNet~\cite{hao2022real} exhibits notable speed but lacks accuracy. In contrast, our HAFormer has achieved a better balance between these aspects. The lower overall performance on the CamVid dataset, compared to Cityscapes, is due to its smaller size and lower resolution, which highlights the robust generalization capability of our approach. Visualization results in Fig.~\ref{fig:camd} further demonstrate the advantages of our HAFormer.

\textbf{Speed Comparison.}
To ensure a fair comparison, all methods are executed on the same platform, as the computational load directly impacts inference speed, which can vary depending on the device. In our controlled evaluation, a single NVIDIA RTX 2080Ti GPU is utilized to measure model execution times. The comparison of speed and runtime between our proposed HAFormer and other lightweight methods is detailed in Table~\ref{tab:efficiency}. The experiments involve the spatial resolution of $512 \times 1024$ for evaluation, aligning with methods with the official code to ensure fairness. Table~\ref{tab:efficiency} demonstrates the impressive speed of HAFormer, achieving a frame rate of 105 fps when processing image streams of size $512 \times 1024$, positioning it as one of the fastest methods. While DABNet operates at 139 fps, HAFormer's competitive accuracy of 74.2\% is significant for real-world applications like autonomous driving. Balancing speed (105 fps) and accuracy effectively, HAFormer emerges as a strong candidate for practical use.

\section{Conclusions}
\label {sec5}

In this study, we introduced HAFormer, a new lightweight semantic segmentation approach. We designed the Hierarchy-Aware Pixel-Excitation Module (HAPE) to extract enhanced hierarchical local features. Additionally, an Efficient Transformer module efficiently captures extensive global features with a limited computational load. Then, we incorporated a correlation-weighted Fusion (cwF) mechanism to combine highly correlated CNN and Transformer features for improved representational learning. Through extensive experiments on benchmark datasets, our approach has shown effectiveness and generalization, highlighting the capability of HAFormer to achieve a balanced trade-off between segmentation accuracy and computational efficiency.

\bibliographystyle{IEEEtran}
\bibliography{reference}


\begin{IEEEbiography}
[{\includegraphics[width=1in,height=1.25in,clip,keepaspectratio]{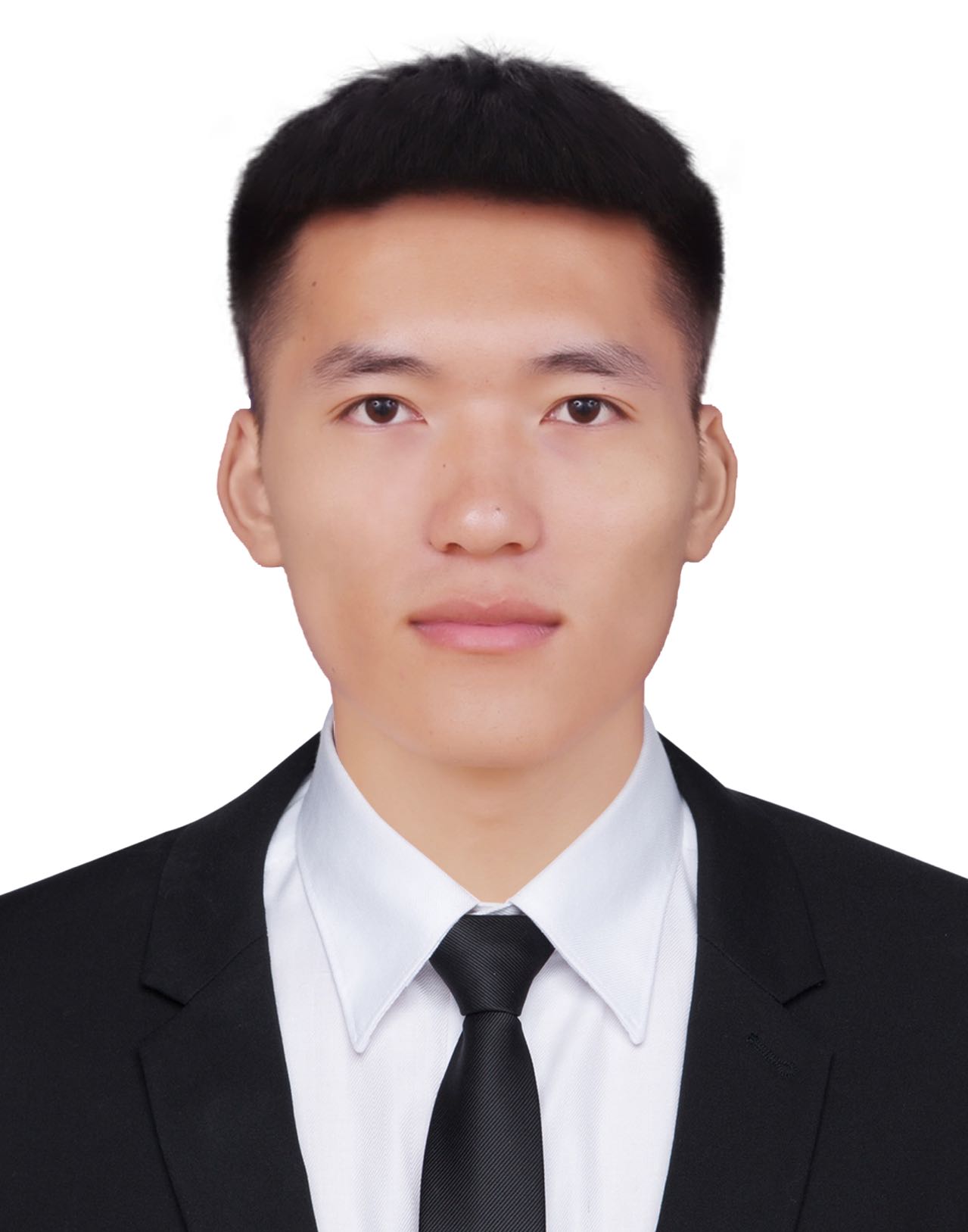}}]
{Guoan Xu}
received the M.S. degree with the College of Automation and College of Artificial Intelligence, Nanjing University of Posts and Telecommunications. He is currently pursuing the Ph.D. degree with the Faculty of Engineering and Information Technology (FEIT), University of Technology Sydney (UTS). His main research interests include image segmentation, and multi-modality image processing.
\end{IEEEbiography}

\begin{IEEEbiography}[{\includegraphics[width=1in,height=1.25in,clip,keepaspectratio]{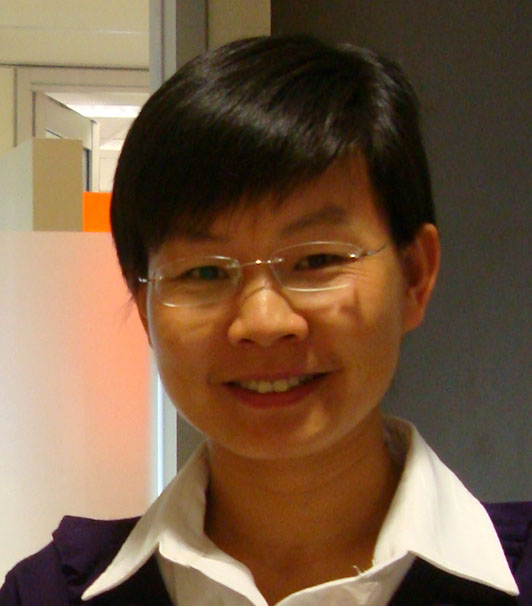}}]{Wenjing Jia} received her Ph.D. degree in Computing Sciences from the University of Technology Sydney in 2007. She is currently an Associate Professor at the Faculty of Engineering and Information Technology (FEIT), University of Technology Sydney (UTS). Her research falls in the fields of image processing and analysis, computer vision, and pattern recognition.
\end{IEEEbiography}

\begin{IEEEbiography}
[{\includegraphics[width=1in,height=1.25in,clip,keepaspectratio]{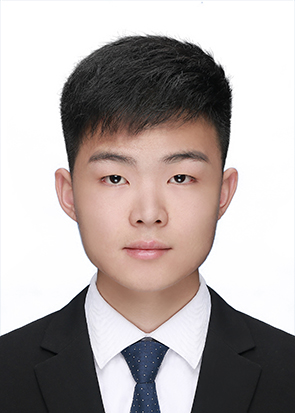}}]
{Tao Wu}
received the BSc degree in Computer Science and Technology from Jilin University, Changchun, China, in 2020. He is currently working toward the Ph.D. degree with the Department of Computer Science and Technology, Nanjing University. His research interests include computer vision and deep learning.
\end{IEEEbiography}

\begin{IEEEbiography}
[{\includegraphics[width=1in,height=1.25in,clip,keepaspectratio]{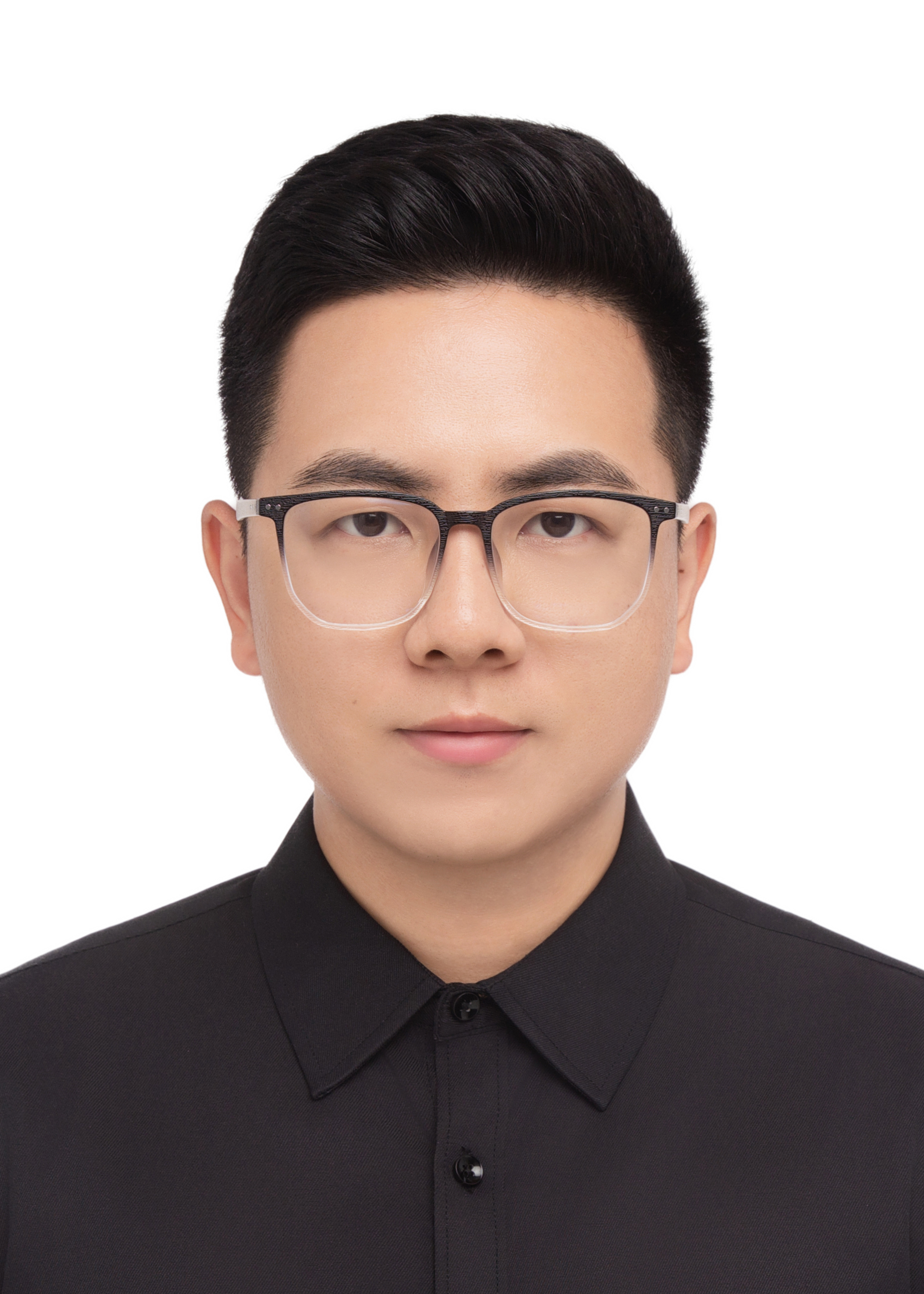}}]
{Ligeng Chen}
received the BSc degree in Computer Science and Technology from Chongqing University, Chongqing, China, in 2017.  He received his Ph.D. degree from the Department of Computer Science and Technology, Nanjing University, Jiangsu, China, in 2023. He is currently a researcher and software engineer in Honor Device Co., Ltd. His research interests include data mining and binary code analysis.
\end{IEEEbiography}

\begin{IEEEbiography}
[{\includegraphics[width=1in,height=1.25in,clip,keepaspectratio]{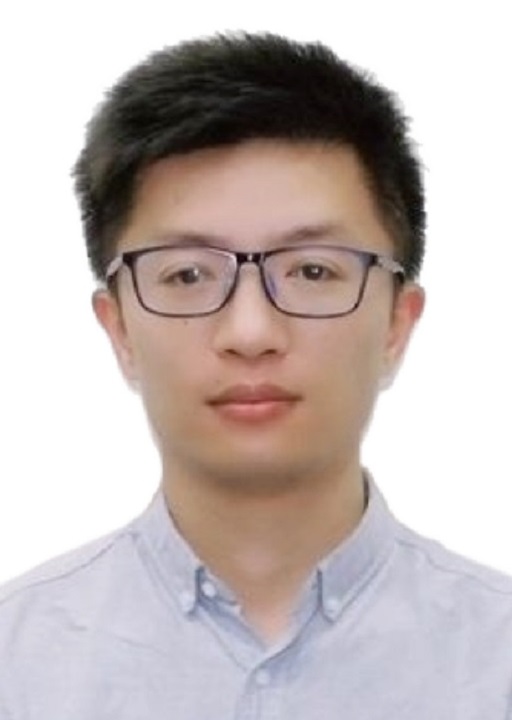}}]
{Guangwei Gao}
(Senior Member, IEEE) received the Ph.D. degree in pattern recognition and intelligence systems from the Nanjing University of Science and Technology, Nanjing, in 2014. He is currently an Associate Professor in Nanjing University of Posts and Telecommunications. His research interests include pattern recognition and computer vision. He has published more than 70 scientific papers in IEEE TIP, IEEE TCSVT, IEEE TITS, IEEE TMM, PR, AAAI, IJCAI, etc. Personal website: \textit{https://guangweigao.github.io}.
\end{IEEEbiography}


\end{document}